\pdfoutput=1

\documentclass[11pt]{article}

\usepackage[final]{acl}

\usepackage{times}
\usepackage{latexsym}

\usepackage[T1]{fontenc}

\usepackage[utf8]{inputenc}

\usepackage{microtype}

\usepackage{inconsolata}

\usepackage{graphicx}

\usepackage{adjustbox}
\usepackage{multirow}
\usepackage{multicol}
\usepackage{booktabs}
\usepackage{setspace}
\usepackage{hyperref}
\usepackage{threeparttable}
\usepackage{xcolor}
\usepackage{colortbl}
\usepackage{float}
\usepackage{enumitem}
\usepackage{array}
\usepackage{arydshln}
\usepackage{makecell}
\usepackage{amsmath}
\usepackage{subfigure}
\usepackage{dashrule}
\usepackage{xspace}
\newcommand{\method}{{MathFusion}}
\newcommand{\dataset}{{MathFusionQA}}
\newcommand{\pa}{{$P_A$}}
\newcommand{\pb}{{$P_B$}}
\newcommand{\pf}{{$P_F$}}
\newcommand{\pfseq}{{$P_F^{\text{seq}}$}}
\newcommand{\pfpara}{{$P_F^{\text{para}}$}}
\newcommand{\pfcond}{{$P_F^{\text{cond}}$}}

\def\huggingface{\raisebox{-1.5pt}{\includegraphics[height=1.05em]{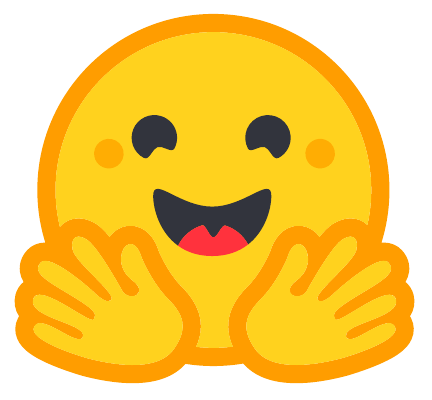}}}
\def\github{\raisebox{-1.5pt}{\includegraphics[height=1.05em]{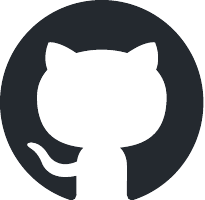}}}
\newcommand{\hflink}{https://huggingface.co/collections/QizhiPei/mathfusion-67d92b8e505635db1baf20bb}
\newcommand{\ghlink}{https://github.com/QizhiPei/MathFusion}

\usepackage{algorithm}
\usepackage{algpseudocode}
\usepackage{colortbl}

\usepackage{amssymb}
\usepackage{mathtools}
\usepackage{amsthm}

\usepackage{colortbl}
\usepackage{nicematrix}

\usepackage{amsthm}
\usepackage{amsmath,amsfonts,bm}
\usepackage{nicefrac}
\usepackage{array}

\usepackage{url}
\usepackage{pbox}

\definecolor{bluex}{rgb}{0.27, 0.42, 0.81}
\definecolor{purplex}{HTML}{9564bf}
\definecolor{red3}{HTML}{C52A20}
\definecolor{red2}{HTML}{B36A6F}
\definecolor{red1}{HTML}{FFb5b5}
\definecolor{purple}{HTML}{B36A6F}
\definecolor{darkyellow}{HTML}{D5BA82}
\definecolor{blue1}{HTML}{508AB2}
\definecolor{blue2}{HTML}{C4E4E3}
\definecolor{green1}{HTML}{A1D0C7}
\definecolor{green2}{HTML}{BFF6BA}
\definecolor{green3}{HTML}{028100}
\definecolor{teal}{HTML}{508AB2}
\definecolor{purple1}{HTML}{8d3a94}

\usepackage{pifont}
\newcommand{\cmark}{\text{\ding{51}}}
\newcommand{\xmark}{\text{\ding{55}}}

\usepackage[listings]{tcolorbox}
\tcbuselibrary{listings,theorems}
\newtcolorbox{mybox}{colback=white!5!white,colframe=black!75!black, left=.05in, right=.05in}

\newtcbtheorem[number within=section]{exmp}{Example}%
{beforeafter skip balanced=.01in,colback=green2!5,colframe=blue1,fonttitle=\small \bfseries, left=.02in, right=.02in,bottom=.02in, top=.02in}{exmp}
\newtcbtheorem[]{trainprompt}{Prompt}%
{colback=green2!5,colframe=blue1,fonttitle=\bfseries, left=.02in, right=.02in,bottom=.02in, top=.02in}{trainprompt}

\theoremstyle{plain}

\theoremstyle{definition}

\theoremstyle{remark}

\title{MathFusion: Enhancing Mathematical Problem-solving of LLM\\ through Instruction Fusion}

\author{
    Qizhi Pei\textsuperscript{1,2},
    {\bf Lijun Wu\textsuperscript{2}$^{\ast}$},
    Zhuoshi Pan\textsuperscript{2,3},
    Yu Li\textsuperscript{2},
    Honglin Lin\textsuperscript{2},\\
    {\bf Chenlin Ming\textsuperscript{2,4}},
    {\bf Xin Gao\textsuperscript{2}},
    {\bf Conghui He\textsuperscript{2}$^{\ast}$},
    {\bf Rui Yan\textsuperscript{1,5,6}\thanks{\ \ Corresponding authors: Lijun Wu (\url{wulijun@pjlab.org.cn}), Conghui He (\url{heconghui@pjlab.org.cn}), and \mbox{Rui Yan} (\url{ruiyan@ruc.edu.cn}).}} \\
    \textsuperscript{1}Gaoling School of Artificial Intelligence, Renmin University of China \\
    \textsuperscript{2}Shanghai Artificial Intelligence Laboratory \quad
    \textsuperscript{3}Tsinghua University \quad
    \textsuperscript{4}Shanghai Jiao Tong University \\
    \textsuperscript{5}Engineering Research Center of Next-Generation Intelligent Search and Recommendation, MOE \\
    \textsuperscript{6}School of Artificial Intelligence, Wuhan University \\
    \texttt{\{qizhipei,ruiyan\}@ruc.edu.cn} \quad
    \texttt{\{wulijun,heconghui\}@pjlab.org.cn} \\
    \textbf{\huggingface} \xspace \href{\hflink} {\texttt{https://huggingface.co/collections/QizhiPei/MathFusion}} \\ \textbf{\github} \xspace \url{\ghlink} \\
}

\begin{document}
\maketitle
\begin{abstract}
Large Language Models (LLMs) have shown impressive progress in mathematical reasoning. While data augmentation is promising to enhance mathematical problem-solving ability, current approaches are predominantly limited to instance-level modifications—such as rephrasing or generating syntactic variations—which fail to capture and leverage the intrinsic relational structures inherent in mathematical knowledge.
Inspired by human learning processes, where mathematical proficiency develops through systematic exposure to interconnected concepts, we introduce \textbf{\mbox{MathFusion}}, a novel framework that enhances mathematical reasoning through cross-problem instruction synthesis.
MathFusion implements this through three fusion strategies: (1) \textit{sequential fusion}, which chains related problems to model solution dependencies; (2) \textit{parallel fusion}, which combines analogous problems to reinforce conceptual understanding; and (3) \textit{conditional fusion}, which creates context-aware selective problems to enhance reasoning flexibility.
By applying these strategies, we generate a new dataset, \textbf{MathFusionQA}, followed by fine-tuning models (DeepSeekMath-7B, Mistral-7B, Llama3-8B) on it.
Experimental results demonstrate that MathFusion achieves substantial improvements in mathematical reasoning while maintaining high data efficiency, boosting performance by 18.0 points in accuracy across diverse benchmarks while requiring only 45K additional synthetic instructions, representing a substantial improvement over traditional single-instruction approaches.
\end{abstract}

\section{Introduction}
Large Language Models (LLMs) have demonstrated remarkable capabilities in various reasoning tasks~\citep{cot_reasoning,llm_reasoning_survey}, with mathematical problem-solving emerging as a critical domain for assessing their cognitive abilities~\citep{llm_math_survey}.
\begin{figure}
  \centering
  \includegraphics[width=0.97\linewidth]{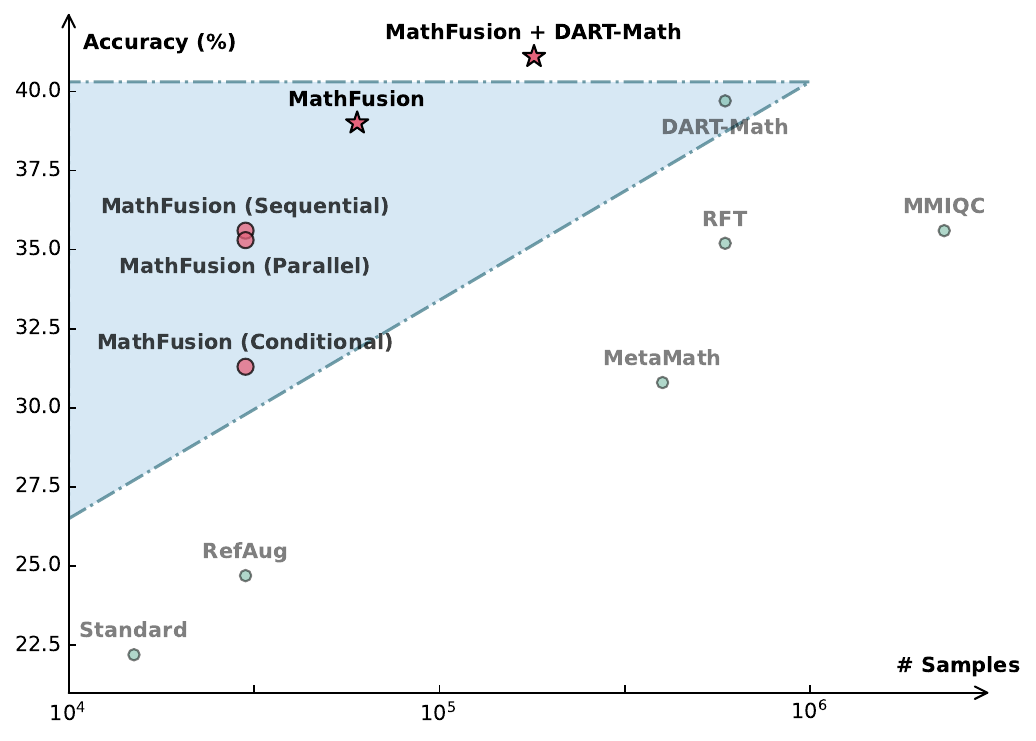}
  \caption{Average performance across six benchmarks of mathematical LLMs built on Llama3-8B, along with the respective \# SFT samples.
  \method~yields superior performance with fewer synthetic instructions.}
  \label{fig:general_performance}
  \vspace{-0.5cm}
\end{figure}
Specialized mathematical LLMs have emerged to address the unique challenges of solving complex mathematical problems~\citep{qwen25_math,deepseekmath,internlm_math,mathstral}.
Current approaches to enhance mathematical reasoning primarily focus on four paradigms: continued pre-training with math corpora~\citep{qwen25_math,deepseekmath}, reinforcement learning (RL) from human or automated feedback~\citep{wizardmath,step_control_dpo}, test-time compute scaling~\citep{mathcoder,mindstar,rstar_math,mathcritique}, and supervised fine-tuning (SFT) using problem-solution pairs~\citep{mathscale,dartmath,lin2025metaladder}.
Among these, SFT is the most widely adopted paradigm~\citep{setlur2024rl} due to its simplicity.
However, its effectiveness is often limited by the complexity and diversity of the mathematical training data~\citep{wizardmath} during SFT.
To this end, data augmentation and synthesis have emerged as promising directions to enhance mathematical reasoning. For example, approaches such as MetaMath~\citep{metamath} and WizardMath~\citep{wizardmath} emphasize enhancing individual problems through rephrasing and difficulty variation. 

While instance-level modifications have shown potential, they do not resolve the fundamental challenge: the inability of LLMs to effectively capture and leverage the intrinsic relational structures that characterize mathematical knowledge~\citep{chu2024beyond,srivatsa2024makes}.
This limitation becomes particularly apparent in real-world scenarios, where complex mathematical problems are often composed of interdependent sub-problems that form intricate dependency graphs~\citep{bagherzadeh2019problem,prabawa2023problem}. 
For instance, solving a system of equations requires the sequential solution of individual equations, followed by the reconciliation of constraints.

Motivated by the way human learners develop proficiency through systematic exposure to interconnected ideas~\citep{komarudin2021analysis}, we propose \textbf{\method}, a novel framework that enhances mathematical reasoning by fusing different mathematical problems. The key insight behind \method~is that the strategic combination of complementary mathematical instructions can unlock deeper reasoning capabilities.
Specifically, by combining two existing problems, \method~synthesizes a new math problem that encapsulates the relational and compositional aspects of the original two problems.
To achieve this, we introduce three distinct fusion strategies:
(1) \textbf{\textit{sequential fusion}}, which links related problems by chaining them together through shared variables to model solution dependencies; 
(2) \textbf{\textit{parallel fusion}}, which integrates analogy problems to enhance conceptual comprehension and generates a novel problem that encapsulates their shared mathematical essence; and
(3) \textbf{\textit{conditional fusion}}, which generates selective problems based on specific context to promote flexible reasoning.

Starting from existing datasets, we first identify pairs of problems that are suitable for fusion.
Then we generate new problems by applying these fusion strategies to pairs of mathematical problems that share similar types and contexts.
After that, we use strong LLMs to generate corresponding solutions. The resulting dataset, \textbf{\dataset}, is then used to fine-tune LLMs including DeepSeekMath-7B, Mistral-7B, and Llama3-8B.

Experimental results demonstrate that \method~enables LLMs to effectively capture the underlying relational structures of mathematical tasks, thereby enhancing their capacity to resolve complex, multi-step problems. Moreover, \method~yields considerable improvements in mathematical reasoning accuracy across both in-domain and challenging out-of-domain benchmarks, outperforming traditional single-instruction fine-tuning by 18.0 points in accuracy on average while incorporating only 45K additional synthetic instructions.
Further integration with the state-of-the-art (SOTA) data augmentation method DART-Math~\citep{dartmath} and scaling \method~to larger size lead to additional improvements, surpassing DART-Math in accuracy on average while utilizing less than one-third of its data.
This highlights the complementary nature and scalability of our approach.

\begin{figure*}
  \vspace{-0.7cm}
  \centering
  \includegraphics[width=\linewidth]{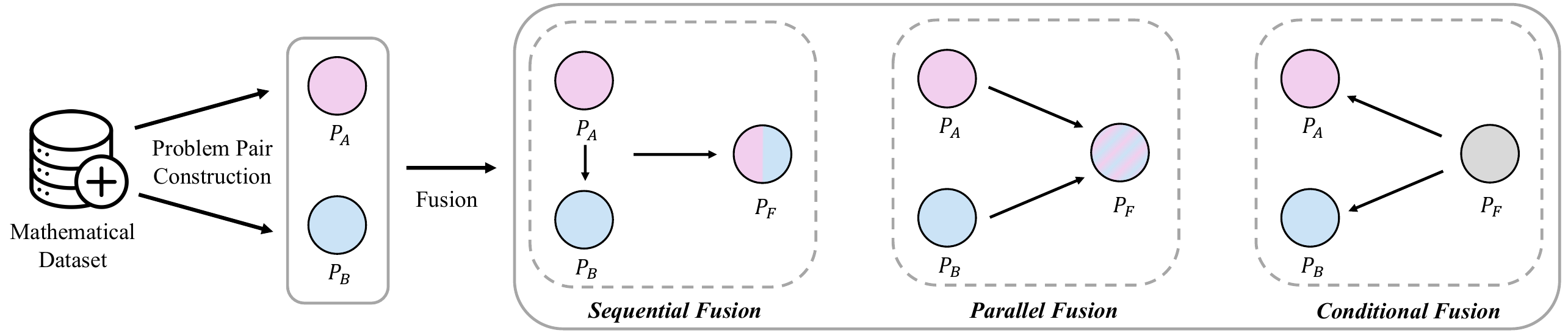}
  \caption{The overview of \method. Given two mathematical problems \pa~and \pb~from the original mathematical dataset, \method~synthesizes a new mathematical problem \pf~by fusing these two problems through three fusion strategies: \textit{sequential fusion}, \textit{parallel fusion}, and \textit{conditional fusion}.}
  \label{fig:overview}
\end{figure*}
\section{Related Work}
\subsection{Individual Data Augmentation for Math}
Existing mathematical data augmentation methods primarily focus on two aspects: \textit{enhancing existing data} and \textit{generating new data}.
\textit{Enhancing existing data} typically involves modifying the problem or solution.
For the problem, strategies include altering the level of complexity/difficulty~\citep{wizardmath}, rephrasing the wording~\citep{metamath,muggle_math}, and employing backward reasoning~\citep{metamath}.
For the solution, methods such as generating diverse and high-quality mathematical reasoning paths through multiple calls~\citep{metamath,muggle_math,refaug,dartmath}, and incorporating reflection~\citep{refaug,pan2025lemma} are commonly used.
\textit{Generating new data} typically involves creating new mathematical problems based on key mathematical concepts~\citep{mathscale}, seed datasets~\citep{scale_quest}, specific example~\citep{xwin_math}, and then using strong mathematical models~\citep{openai2023gpt4,deepseekmath} to generate corresponding solutions.
These methods, however, focus primarily on individual mathematical problems, overlooking the underlying relationships between different problems.

\subsection{Compositional Data Augmentation}
Most data augmentation methods focus on enhancing individual instances, while few consider the relationships between different instances.
\textit{mixup}~\citep{mixup} is an augmentation technique that addresses this gap by generating synthetic training samples through linear interpolations between pairs of input data points and their corresponding labels, which has been shown to be effective across various tasks~\citep{mixup_survey1,mixup_survey2}, such as image classification~\citep{mixup,mixup_train}, text classification~\citep{guo2019augmenting,zhang2020seqmix}, and neural machine translation~\citep{guo2020sequence,wu2021mixseq}.
Mosaic-IT~\citep{mosaic} is a model-free data augmentation method that concatenates instruction data and trains LLMs with meta-instructions, thereby enhancing performance and reducing training costs.
Some works also consider the composition of multiple skills or keypoints. 
Instruct-SkillMix~\citep{instruct_skillmix} extracts core skills for instruction-following and generates new instructions by randomly combining pairs of skills.
KPMath~\citep{kpmath} shares the same idea with Instruct-SkillMix, but focuses on mathematical problems by extracting topics and key points from the problem and generates new problems by combining them.

In contrast to existing works, our approach primarily focuses on fusing mathematical problems and places particular emphasis on the logical coherence of the fusion.

\section{\method}
The overview of \method~is shown in Figure~\ref{fig:overview}. 
Given two mathematical problems \pa~and \pb~from the original mathematical training set, \method~synthesizes a new mathematical problem \pf~by fusing these two problems.
A simple example for \pa~and \pb~is shown in Example~\ref{exmp:main_original_questions}, and we show the corresponding \pf~for three fusion strategies in the following sections.
More cases are shown in the Appendix~\ref{sec:appendix_more_cases}.
\vspace{2mm}
\begin{exmp}{Original Questions}{main_original_questions}
\small
\pa: During one day, there are 4 boat trips through the lake. The boat can take up to 12 people during one trip. How many people can the boat transport in 2 days?\\\\
\pb: The school is organizing a trip to the museum. 4 buses were hired to take the children and teachers to their destination. The second bus has twice the number of people on it as the first bus. The third bus has 6 fewer people than the second bus. The fourth bus has 9 more people than the first bus. If the first bus has 12 people, how many people are going to the museum in total?
\end{exmp}
\vspace{2.5mm}
In the following sections, we will first introduce the problem pair construction in Section~\ref{sec:problem_pair}, and then introduce the three fusion strategies: \textit{sequential fusion} in Section~\ref{sec:sequential_fusion}, \textit{parallel fusion} in Section~\ref{sec:parallel_fusion}, and \textit{conditional fusion} in Section~\ref{sec:conditional_fusion}.
Based on the augmented problem sets generated by these fusion strategies, we present the \dataset~dataset in Section~\ref{sec:mathfusionqa}.

\subsection{Problem Pair Construction}
\label{sec:problem_pair}
To construct problem pairs for fusion, for each problem \pa, we need to identify a suitable problem \pb~. A straightforward approach is to select a problem \pb~that shares the same type and similar context with \pa. 
Formally, the problem pair set $\mathcal{D}_{\text{pair}}^{\text{p}}$ is defined as:
\begin{equation*}
  \scriptsize
  \begin{gathered}
    \mathcal{D}_{\text{pair}}^{\text{p}} = \left\{(P_A, P_B) \mid P_A \in \mathcal{D}_{\text{train}}^{\text{p}}, P_B = \underset{P \in \mathcal{D}_{\text{train}}^{\text{p}} \setminus \{P_A\}}{\arg\max} \text{SIM}(P_A, P) \right\},
  \end{gathered}
\end{equation*}
where $\mathcal{D}_{\text{train}}^{\text{p}}$ is a set of problems from the original training set, and $\text{SIM}(P_A, P_B)$ is the inner product of the embeddings of \pa~and \pb~using OpenAI embedding API \textit{text-embedding-3-large}~\citep{openai2023gpt4}.

\subsection{Sequential Fusion}
\label{sec:sequential_fusion}
In mathematical problem-solving, sequential reasoning is a common pattern where the solution of the whole problem is the sequential combination of the solutions of the sub-problems. 
\textit{Sequential fusion} constructs a new mathematical problem \pfseq~by establishing solution dependencies between two original problems \pa~and \pb~through shared variables, where the answer of \pa~becomes a prerequisite for solving \pb.
Formally, the \textit{sequential fusion} process and the resulting augmented problem set are defined as: 
\begin{equation*}
  \small
  \begin{gathered}
  P_{F}^{\text{seq}} = P_B(P_A),
  \mathcal{D}_{\text{seq}}^{\text{p}} = \{ P_{F}^{\text{seq}} \mid (P_A, P_B) \in \mathcal{D}_{\text{pair}}^{\text{p}}\}.
  \end{gathered}
\end{equation*}
The answer from solving $P_A$ serves as a part of the input to $P_B$, thereby creating a chained dependency. A specific example of \textit{sequential fusion} is shown in Example~\ref{exmp:main_sequential}. The answer of \pa~(the number of people transported by the boat) is used as the input for \pb~(the number of people in the first bus).
\begin{exmp}{Sequential Fusion}{main_sequential}
  \small
  $P_{F}^{\text{seq}}$: The school has organized a trip to a museum and needs to transport children and teachers. \textcolor{red3}{First, calculate how many people can be transported} by a boat over 2 days, with 4 boat trips each day, and each trip can carry up to 12 people. \textcolor{red3}{Let this total be the number of people in the first bus.} The second bus has twice the number of people on the first bus, the third bus has 6 fewer people than the second bus, and the fourth bus has 9 more people than the first bus. \textcolor{red3}{How many people are going to the museum in total?}
  \end{exmp}
\vspace{-0.2cm}

\subsection{Parallel Fusion}
\label{sec:parallel_fusion}
Analogous problems often share common mathematical concepts and essences. 
\textit{Parallel fusion} leverages this by synthesizing \pfpara~through the integration of two conceptually analogous problems \pa~and \pb~, thereby creating a new problem that encapsulates their shared mathematical essence. 
This approach emphasizes the conceptual relationships between problems rather than their sequential dependencies.
The \textit{parallel fusion} process and the resulting augmented problem set are formally defined as:  
\begin{equation*}
  \small
  \begin{gathered}
    P_A \rightarrow P_A', P_B \rightarrow P_B', P_{F}^{\text{para}} = \Phi(P_A', P_B'), \\
    \mathcal{D}_{\text{para}}^{\text{p}} = \{ P_{F}^{\text{para}} \mid (P_A, P_B) \in \mathcal{D}_{\text{pair}}^{\text{p}}\},
  \end{gathered}
\end{equation*}
where $P_A'$ and $P_B'$ denote the potentially modified problems from \pa~and \pb, respectively, for the fused problem \pfpara. The function $\Phi$ encompasses various operations, such as algebraic composition and the enforcement of constraint satisfiability, to rigorously integrate the underlying mathematical structures.
A concrete illustration of \textit{parallel fusion} is provided in Example~\ref{exmp:main_parallel}.
The total number of people transported by boat and buses over 2 days is asked to be calculated, and the input of $P_A'$(the number of trips made by the boat in one day) is different from that of \pa.
\begin{exmp}{Parallel Fusion}{main_parallel}
\small
$P_{F}^{\text{para}}$: \textcolor{red3}{A school organizes a field trip to a museum and hires 4 buses and a boat}. The boat makes \textcolor{blue}{2} trips in one day, with a capacity of 12 people per trip. Each bus has a different number of people: the first bus bus has 12 people ... the fourth bus has 9 more people than the first bus.

\textcolor{red3}{Calculate the total number of people transported by the boat and the buses over the course of 2 days. How many people can the boat and buses transport in total for the trip?}
\end{exmp}
\vspace{2.5mm}

\subsection{Conditional Fusion}
\label{sec:conditional_fusion}  
Context-aware reasoning necessitates the dynamic selection or comparison of solutions based on conditional constraints. \textit{Conditional fusion} synthesizes \pfcond~by integrating \pa~and \pb~into a cohesive real-world scenario, where the final solution is derived through contextual comparison or selection of outcomes from \pa~and \pb.
Formally, the \textit{conditional fusion} process and the resulting augmented problem set are defined as:
\begin{equation*}
  \small
  \begin{gathered}
  P_{F}^{\text{cond}} = \Gamma(P_A, P_B),
  \mathcal{D}_{\text{cond}}^{\text{p}} = \{ P_{F}^{\text{cond}} \mid (P_A, P_B) \in \mathcal{D}_{\text{pair}}^{\text{p}}\}.
  \end{gathered}
\end{equation*}  
$\Gamma$ is a comparison function that contrasts $P_A$ and $P_B$ based on predefined logical or contextual rules.
A concrete case is shown in Example~\ref{exmp:main_conditional}, where the final solution is determined by comparing the answers of \pa~(the capacity of the boat) and \pb~(the capacity of the buses) in a real-world scenario (organizing a lake excursion and a museum trip).
\begin{exmp}{Conditional Fusion}{main_conditional}
\small
$P_{F}^{\text{cond}}$: \textcolor{red3}{A local community is organizing two different outings}. \textcolor{red3}{For a lake excursion}, a boat operates 4 trips a day with a capacity of 12 people per trip. They plan to run this boat service for 2 days. \textcolor{red3}{Meanwhile, a school is arranging a trip to the museum} with 4 buses. The first bus has 12 people, the second bus has twice as many people as the first, the third bus has 6 fewer people than the second, and the fourth bus has 9 more people than the first bus. \textcolor{red3}{Given these arrangements, which mode of transportation has a larger capacity for transporting people?}
\end{exmp}
\vspace{2.5mm}
To clarify, the core difference between \textit{parallel fusion} and \textit{conditional fusion} is that: \textit{parallel fusion} combines \pa~and \pb~to form a novel \pfpara, where the input of \pfpara~may be different from the original \pa~and \pb; while \textit{conditional fusion} compares the results of \pa~and \pb, the input of \pfcond~is the same as \pa~and \pb, and the output is based on the comparison of the results of \pa~and \pb.

\subsection{\dataset~Dataset}
\label{sec:mathfusionqa}
After applying the three fusion strategies to $\mathcal{D}_{\text{pair}}^{\text{p}}$ and get the augmented problem sets $\mathcal{D}_{\text{seq}}^{\text{p}}$, $\mathcal{D}_{\text{para}}^{\text{p}}$, and $\mathcal{D}_{\text{cond}}^{\text{p}}$, we use GPT-4o-mini~\citep{openai2023gpt4} to generate corresponding solutions $S$ for the augmented problems.
The resultingaugmented data $\mathcal{D}_{\text{seq}}$, $\mathcal{D}_{\text{para}}$, and $\mathcal{D}_{\text{cond}}$ are combined with the original training set $\mathcal{D}_{\text{train}}$ to form the final \dataset~dataset as $\mathcal{D}_{\text{\dataset}}=\mathcal{D}_{\text{train}} \cup \mathcal{D}_{\text{seq}} \cup \mathcal{D}_{\text{para}} \cup \mathcal{D}_{\text{cond}}$.
We use GSM8K~\citep{cobbe2021gsm8k} and MATH~\citep{hendrycks2021math} as the original training set separately.
We compare our \dataset~with other mathematical datasets in Table~\ref{tab:datasets_stat}.
Though \dataset~ is overall smaller than other datasets except for RefAug~\citep{refaug}, we empirically show that \dataset~exhibits strong performance and is more effective than RefAug in Section~\ref{sec:main_results}.
Then we fine-tune LLMs on the \dataset~dataset, resulting in \method~models.
Given that some problems in \dataset~may be incomplete or incorrect, we conduct an analysis of problem evaluation and error correction, and present cases of unsuitable fusions in Appendix~\ref{sec:appendix_problem_analysis}.
\begin{table}[t]
    \centering
    \resizebox{\linewidth}{!}{
      \begin{tabular}{lr}
        \toprule
        Dataset            & \# Samples  \\ \midrule
        WizardMath~\citep{wizardmath} & 96K \\
        MetaMathQA~\citep{metamath}         & 395K  \\
        MMIQC~\citep{liu2024mmiqc}              & 2294K      \\
        Orca-Math~\citep{mitra2024orcamath} & 200K \\
        Xwin-Math-V1.1~\citep{xwin_math}  & 1440K \\
        KPMath-Plus~\citep{kpmath}  & 1576K \\
        MathScaleQA~\citep{mathscale} & 2021K \\
        DART-Math-Uniform~\citep{dartmath}  & 591K  \\
        DART-Math-Hard~\citep{dartmath}  & 585K  \\
        RefAug~\citep{refaug}  & 30K \\
        \midrule
        \dataset & 60K \\
        \bottomrule
    \end{tabular}}
    \caption{Comparison between \dataset~and previous mathematical datasets. 
    Our \dataset~is generally smaller than others.
    }
    \label{tab:datasets_stat}
  \end{table}

\section{Experiments}
\subsection{Experimental Setup}
\noindent{\textbf{Data Synthesis:}}
We use GPT-4o-mini~\citep{openai2023gpt4} to fuse the problems and generate the corresponding solutions.
The details about generation and corresponding prompts are shown in Appendix~\ref{sec:appendix_synthesis_setup} and~\ref{sec:appendix_prompts}).

\noindent{\textbf{Training:}}
We conduct standard instruction-tuning on our \dataset.
Following DART-Math~\citep{dartmath}, we conduct experiments on two categories of base models: 7B math-specialized base LLM, specifically DeepSeekMath-7B~\citep{deepseekmath}, and 7-8B general base LLMs, specifically Mistral-7B~\citep{jiang2023mistral} and Llama3-8B~\citep{llama3}.
Each base model is fine-tuned using three distinct fusion strategies: \textit{sequential}, \textit{parallel}, and \textit{conditional}.
For each strategy, the fine-tuning dataset (30K samples) comprises the union of the GSM8K, MATH datasets, and an augmented set generated by the respective fusion strategy.
The \dataset{} dataset (60K samples) is formed by the union of all these sub-datasets.
Table~\ref{tab:mathfusionqa_stat} shows the statistics of the \dataset~collection.
To demonstrate the scaling ability of \method, we also enlarge \dataset~dataset by including top-2 to top-4 nearest neighbors, resulting in a total of 15K + 4 × (3 × 15K) = 195K examples.
All models are trained for 3 epochs.
More details about the training setup are provided in Appendix~\ref{sec:appendix_training_setup}.
\begin{table*}[!htbp]
    \centering
        \resizebox{\textwidth}{!}{
        \begin{tabular}{lccccccccc}
          \toprule
          \multicolumn{1}{c}{\multirow{2}{*}{Model}}                &
          \multirow{2}{*}{\# Samples} & \multicolumn{2}{c}{In-Domain}
          & \multicolumn{4}{c}{Out-of-Domain}
          &              \\
          \cmidrule(lr){3-4} \cmidrule(lr){5-8}
          \multicolumn{1}{c}{}
          &      & MATH                  & GSM8K                 &
          College               & DM                    & Olympiad
          & Theorem               & AVG                   \\     \midrule
          \multicolumn{10}{c}{\textbf{DeepSeekMath (7B Math-Specialized Base Model)}} \\
          \midrule
          DeepSeekMath-7B-RFT
          & 590K & 53.0                  & \textbf{88.2}         &
          \textbf{41.9}         & 60.2                  & 19.1
          & 27.2                  & 48.3                  \\
        DeepSeekMath-7B-DART-Math & 590K & 53.6 & 86.8
        & 40.7        & 61.6 &
        21.7 & \textbf{32.2}          & 49.4 \\
        DeepSeekMath-7B-Instruct
        & 780K & 46.9                  & 82.7                  & 37.1
        & 52.2                  & 14.2                  & 28.1
        & 43.5                  \\
        DeepSeekMath-7B-MMIQC
        & 2.3M  & 45.3                  & 79.0                  & 35.3
        & 52.9                  & 13.0                  & 23.4
        & 41.5                  \\
        \rowcolor[rgb]{ .867, .922, .969} {\method-DSMath-7B} & 195K & \textbf{58.2} & 79.5 & 40.3 & \textbf{69.1} & \textbf{25.5} & 27.0 & \textbf{49.9}\\
        \hdashline
          DeepSeekMath-7B-Standard & 15K & 30.6 & 66.3 & 22.7 & 28.6 & 5.6 & 11.0 & 27.5\\
          DeepSeekMath-7B-RefAug & 30K & 32.1 & 71.2 & 26.0 & 38.4 & 10.1 & 14.4 & 32.0\\
          \rowcolor[rgb]{ .867, .922, .969}  {\method-DSMath-7B} (\textit{Sequential}) & 30K & 49.9 & 76.6 & 38.8 & 64.6 & 21.6 & 22.8 & 45.7\\
          \rowcolor[rgb]{ .867, .922, .969} {\method-DSMath-7B} (\textit{Parallel}) & 30K & 50.9 & 76.7 & 38.9 & 62.2 & 19.0 & 23.8 & 45.3\\
          \rowcolor[rgb]{ .867, .922, .969} {\method-DSMath-7B} (\textit{Conditional}) & 30K & 48.5 & 74.6 & 37.0 & 55.2 & 19.3 & 19.0 & 42.3\\
          DeepSeekMath-7B-MetaMath$^\dagger$ & 60K & 40.0 & 79.0 & 33.2 & 45.9 & 9.5 & 18.9 & 37.8\\
          DeepSeekMath-7B-MMIQC$^\dagger$ & 60K & 26.3 & 60.6 & 19.2 & 41.5 & 10.4 & 6.8 & 27.5 \\
          DeepSeekMath-7B-RefAug$^\dagger$ & 60K & 33.1 & 71.6 & 26.2 & 35.4 & 10.5 & 14.0 & 31.8 \\
            DeepSeekMath-7B-DART-Math$^\dagger$ & 60K & 51.4 & \textbf{82.9} & 39.1 & 62.8 & 21.0 & \textbf{27.4} & 47.4\\
          \rowcolor[rgb]{ .867, .922, .969} {\method-DSMath-7B} & 60K & \textbf{53.4} & 77.9 & \textbf{39.8} & \textbf{65.8} & \textbf{23.3} & 24.6 & \textbf{47.5}\\
          \midrule
          \multicolumn{10}{c}{\textbf{Mistral-7B (7-8B General Base Model)}} \\
          \midrule
  
        Mistral-7B-MetaMath
        & 400K & 29.8                  & 76.5                  & 19.3
        & 28.0                  & 5.9                   & 14.0
        & 28.9                  \\
          Mistral-7B-WizardMath-V1.1
          & 418K    & 32.3                  & 80.4                  & 23.1
          & 38.4                  & 7.7                   & 16.6
          & 33.1                  \\
          Mistral-7B-RFT
          & 590K & 38.7                  & \textbf{82.3}                  & 24.2
          & 35.6                  & 8.7                   & 16.2
          & 34.3                  \\
            Mistral-7B-DART-Math & 590K & \textbf{45.5}          & 81.1
            & \textbf{29.4} & \textbf{45.1} &
            \textbf{14.7} & \textbf{17.0} & \textbf{38.8} \\
            Mistral-7B-MathScale
            & 2.0M  & 35.2                  & 74.8                  & 21.8
            & --                    & --                    & --
            & --                    \\
            Mistral-7B-MMIQC
            & 2.3M  & 37.4                  & 75.4                  & 28.5
            & 38.0                  & 9.4                   & 16.2
            & 34.2                  \\
        \hdashline
          Mistral-7B-Standard & 15K & 12.4 & 60.3 &  8.4 & 17.0 & 2.2 & 7.6 & 18.0\\
          Mistral-7B-RefAug & 30K & 15.1 & 61.1 & 10.4 & 15.4 & 3.1 & 11.0 & 19.4\\
          \rowcolor[rgb]{ .867, .922, .969}  {\method-Mistral-7B} (\textit{Sequential}) & 30K & 32.7 & 73.9 & 18.9 & 29.3 & 9.3 & 15.5 & 29.9\\
          \rowcolor[rgb]{ .867, .922, .969} {\method-Mistral-7B} (\textit{Parallel}) & 30K & 30.9 & 75.1 & 20.9 & 26.5 & 11.0 & 15.2 & 29.9\\
          \rowcolor[rgb]{ .867, .922, .969} {\method-Mistral-7B} (\textit{Conditional}) & 30K & 26.3 & 73.0 & 15.6 & 21.4 & 7.3 & 12.8 & 26.1\\
          Mistral-7B-MetaMath$^\dagger$ & 60K & 22.7 & 70.8 & 14.1 & 27.2 & 5.0 & 12.2 & 25.3 \\
          Mistral-7B-MMIQC$^\dagger$ & 60K & 17.3 & 61.4 & 11.1 & 13.5 & 5.0 & 5.9 & 19.0\\
          Mistral-7B-RefAug$^\dagger$ & 60K & 17.4 & 63.1 & 12.5 & 18.1 & 3.9 & 11.1 & 21.0 \\ 
            Mistral-7B-DART-Math$^\dagger$ & 60K & 34.1 & 77.2 & 23.4 & 36.0 & 8.7 & \textbf{18.2} & 32.9\\
          \rowcolor[rgb]{ .867, .922, .969} {\method-Mistral-7B} & 60K & \textbf{41.6} & \textbf{79.8} & \textbf{24.3} & \textbf{39.2} & \textbf{13.6} & 18.1 & \textbf{36.1} \\
            \midrule
            \multicolumn{10}{c}{\textbf{Llama3-8B (7-8B General Base Model)}} \\
            \midrule
          Llama3-8B-MetaMath
          & 400K & 32.5                  & 77.3                  & 20.6
          & 35.0                  & 5.5                   & 13.8
          & 30.8                  \\
          Llama3-8B-RFT
          & 590K & 39.7                  & \textbf{81.7}                 & 23.9
          & 41.7                  & 9.3                   & 14.9
          & 35.2                  \\
          Llama3-8B-MMIQC
          & 2.3M  & 39.5                  & 77.6                  &
          \textbf{29.5}         & 41.0                  & 9.6
          & 16.2                  & 35.6                  \\
            Llama3-8B-DART-Math & 590K & \textbf{46.6} & 81.1
            & 28.8          & \textbf{48.0}         &
            \textbf{14.5} & \textbf{19.4} & \textbf{39.7} \\
            \hdashline
          Llama3-8B-Standard & 15K & 17.5 & 65.4 & 12.9 & 21.6 & 4.7 & 10.9 & 22.2\\
          Llama3-8B-RefAug & 30K & 20.8 & 67.3 & 15.7 & 25.9 & 4.7 & 13.6 & 24.7\\
          \rowcolor[rgb]{ .867, .922, .969}  {\method-Llama3-8B} (\textit{Sequential}) & 30K & 38.8 & 77.9 & 25.1 & 42.0 & 12.6 & 17.0 & 35.6\\
          \rowcolor[rgb]{ .867, .922, .969} {\method-Llama3-8B} (\textit{Parallel}) & 30K & 38.1 & 75.4 & 25.5 & 41.9 & 11.9 & 18.9 & 35.3\\
          \rowcolor[rgb]{ .867, .922, .969} {\method-Llama3-8B} (\textit{Conditional}) & 30K & 34.7 & 76.9 & 21.2 & 27.4 & 11.9 & 15.5 & 31.3 \\
          Llama3-8B-MetaMath$^\dagger$ & 60K & 28.7 & 78.5 & 19.7 & 31.3 & 5.3 & 16.1 & 29.9\\
          Llama3-8B-MMIQC$^\dagger$ & 60K & 24.4 & 69.7 & 13.4 & 30.9 & 5.2 & 10.6 & 25.7\\
          Llama3-8B-RefAug$^\dagger$ & 60K & 20.3 & 68.6 & 15.5 & 29.1 & 5.5 & 13.0 & 25.3\\
            Llama3-8B-DART-Math$^\dagger$ & 60K & 39.6 & \textbf{82.2} & \textbf{27.9} & 39.9 & 12.9 & \textbf{22.9} & 37.6\\
          \rowcolor[rgb]{ .867, .922, .969} {\method-Llama3-8B} & 60K & \textbf{46.5} & \textbf{79.2} & \textbf{27.9} & \textbf{43.4} & \textbf{17.2} & 20.0 & \textbf{39.0} \\
          \bottomrule
        \end{tabular}
      }
      \caption{
        Performance comparison on mathematical benchmarks including MATH, GSM8K, CollegeMATH (College), DeepMind-Mathematics (DM), OlympiadBench-Math (Olympiad), and TheoremQA (Theorem). 
        The table is organized by the base model and the number of training samples, using 60K as the threshold for splitting.
        The best results are highlighted in bold.
        Rows are sorted according to data size.
        Most of the baseline results are derived from DART-Math~\citep{dartmath}, except for the Standard, RefAug~\citep{refaug}, and baseline labeled with $^\dagger$, which are our own runs.
        \textit{Sequential}, \textit{Parallel}, and \textit{Conditional} indicate training on the union of GSM8K, MATH, and the respective fused dataset.
      }
  \label{tab:main_result}
\end{table*}

\noindent{\textbf{Evaluation:}}
Following DART-Math~\citep{dartmath}, we evaluate the models on two \textbf{in-domain} (ID) benchmarks: GSM8K~\citep{cobbe2021gsm8k} and MATH~\citep{hendrycks2021math}, as our \dataset~dataset is built upon these two datasets.
For \textbf{out-of-domain} (OOD) evaluation, we use the CollegeMath~\citep{mathscale}, DeepMind-Mathematics~\citep{saxton2018dmmath}, OlympiadBench-Math~\citep{he2024olympiadbench}, and TheoremQA~\citep{chen2023theoremqa} benchmarks.
We use greedy decoding to generate solutions for the problems in test sets.
We report the accuracy in 0-shot setting for all models following~\citet{dartmath}.
More details about the evaluation setup and benchmarks are provided in the Appendix~\ref{sec:appendix_evaluation_setup}.

\noindent{\textbf{Baselines:}}
We mainly compare our \method~models with mathematical instruction-based models, which can be categorized into three groups:
(1) Previous top-performing models, including MetaMath~\citep{metamath}, WizardMath~\citep{wizardmath}, RFT (rejection sampling fine-tuning from DART-Math)~\citep{yuan2023rft,dartmath}, MMIQC~\citep{liu2024mmiqc}, MathScale~\citep{mathscale}, DeepSeekMath-7B-Instruct~\citep{deepseekmath}, RefAug~\citep{refaug}, and DART-Math~\citep{dartmath} (we report the \textit{Prop2Diff} version as it generally performs better than the \textit{Uniform} version);
(2) Models instruction-tuned on the combination of GSM8K and MATH datasets (noted as ``Standard'' setting);
(3) Models instruction-tuned on the sampled 60K version of previous top-performing methods to further evaluate the data efficiency of different mathematical data augmentation methods.
Details about the sampling method are introduced in Appendix~\ref{sec:appendix_training_setup}.

\subsection{Main Results}
\label{sec:main_results}
The main results are shown in Table~\ref{tab:main_result}.
We summarize several key findings as follows:

\noindent{\textbf{\textit{Finding 1:} Three fusion strategies consistently enhance the model performance.}}
For all three fusion strategies-\textit{sequential}, \textit{parallel}, and \textit{conditional fusion}—the \method~models consistently surpass the standard settings across all base models and evaluation benchmarks.
Specifically, on MATH and GSM8K test sets, using Llama3-8B as the base model, \method~(\textit{sequential}) achieves 21.3 and 12.5 accuracy improvement; \method~(\textit{parallel}) achieves 20.6 and 10.0 accuracy improvement; and \method~(\textit{conditional}) achieves 18.0 and 11.9 accuracy improvement, respectively, compared to the standard setting.
For four OOD benchmarks, the single fusion strategy also outperforms the standard setting, with a 9.9 accuracy improvement on average.
These improvements demonstrate the effectiveness of the three fusion strategies in enhancing both the ID and OOD generalization performance of the models.

\noindent{\textbf{\textit{Finding 2:} Among three fusion strategies, \textit{sequential fusion} and \textit{parallel fusion} generally perform better than \textit{conditional fusion}.}} A possible reason is that the \textit{conditional fusion} requires no modification of input structures or problem dependencies, merely performing a direct comparison or selection between the solutions of two independent problems without necessitating additional mathematical transformations or reformulations.
We further investigate the difficulty of the problems generated by the three fusion strategies in Section~\ref{sec:problem_difficulty}.

\noindent{\textbf{\textit{Finding 3:} Combination of three fusion strategies further improves performance.}}
As the three fusion strategies capture different aspects of the problem fusion, we further investigate the performance of the combined fusion strategies.
From Table~\ref{tab:main_result}, we observe that the combined fusion strategies consistently outperform each single fusion strategy, indicating that the combination of three fusion strategies can further enhance the model's mathematical ability.
Additionally, the weaker the performance of the base model, the more enhancements the combined fusion strategies can bring.
Specifically, the combined fusion strategies achieve an average accuracy improvement of 3.1 points on DeepSeekMath-7B, 4.9 points on Llama3-8B, and 7.5 points on Mistral-7B across all benchmarks.

\noindent{\textbf{\textit{Finding 4:} Compared with previous top-performing baselines, \method~models yields competitive performance and high data efficiency.}}
For each single fusion strategy, \method~models outperform RefAug, which has the same data size as \method, on all benchmarks.
After combining the three fusion strategies, \method~outperforms previous top-performing baselines like MetaMath and DART-Math on average under the same data size setting.
Specifically, \method~yields consistently better performance on MATH, DeepMind-Mathematics, and OlympiadBench-Math benchmarks.
These results demonstrate the high data efficiency and generalization ability of \method.
\method~maintains also competitive efficacy compared to top-performing models in the full-data regime, exhibiting only a marginal average performance drop on Llama3-8B and DeepSeekMath-7B.

\noindent{\textbf{\textit{Finding 5:} \method~exhibits strong scalability and outperforms larger-scale baselines with fewer samples.}}
The results on DeepSeekMath-7B (the strongest base model for math) in Table~\ref{tab:main_result} reveal that scaling \method~from 60K to 195K samples leads to consistent performance gains across all evaluation benchmarks. Notably, \method-DSMath-7B (195K) surpasses DeepSeekMath-7B-DART-Math (590K) in average accuracy (49.9 vs. 49.4), despite using only one-third of the training data. This illustrates the high scalability and data efficiency of \method. The substantial performance gains on challenging benchmarks such as MATH (+4.6), DeepMind-Mathematics (+7.5), and OlympiadBench-Math (+3.8) underscore the method's capability to generalize well under increased data volume. 
These findings demonstrate that \method~benefits substantially from larger synthetic training sets and can outperform significantly larger instruction-tuned models with less data.

\begin{figure*}[t!]
  \centering
  \subfigure{
    \includegraphics[width=0.31\linewidth]{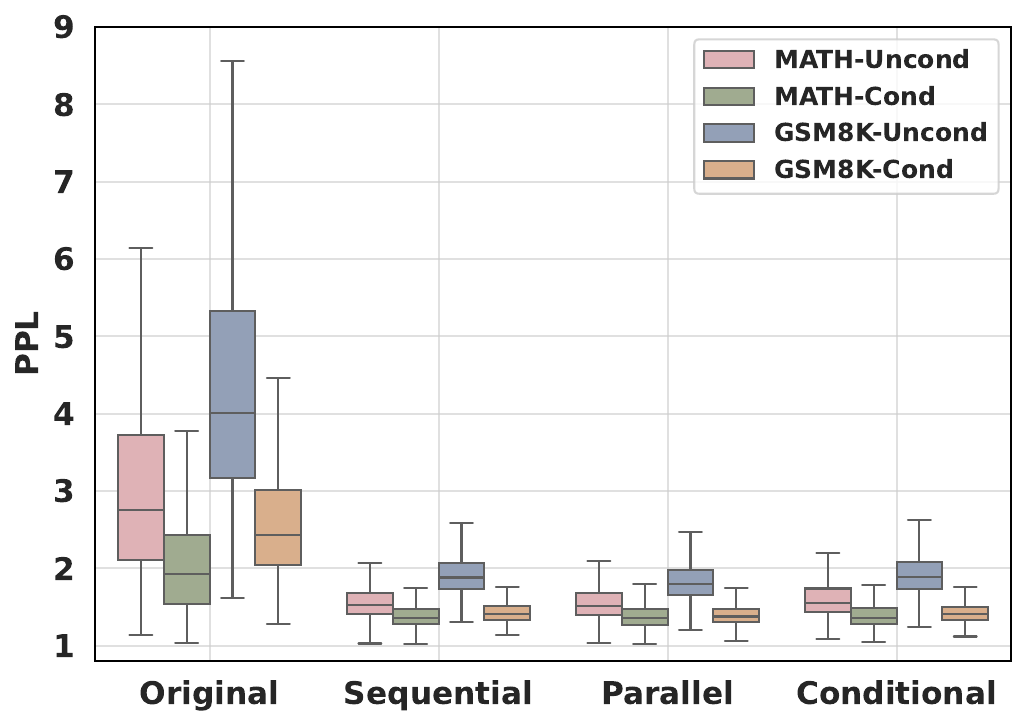}
    \label{fig:analysis_ppl}
  }
  \subfigure{
    \includegraphics[width=0.31\linewidth]{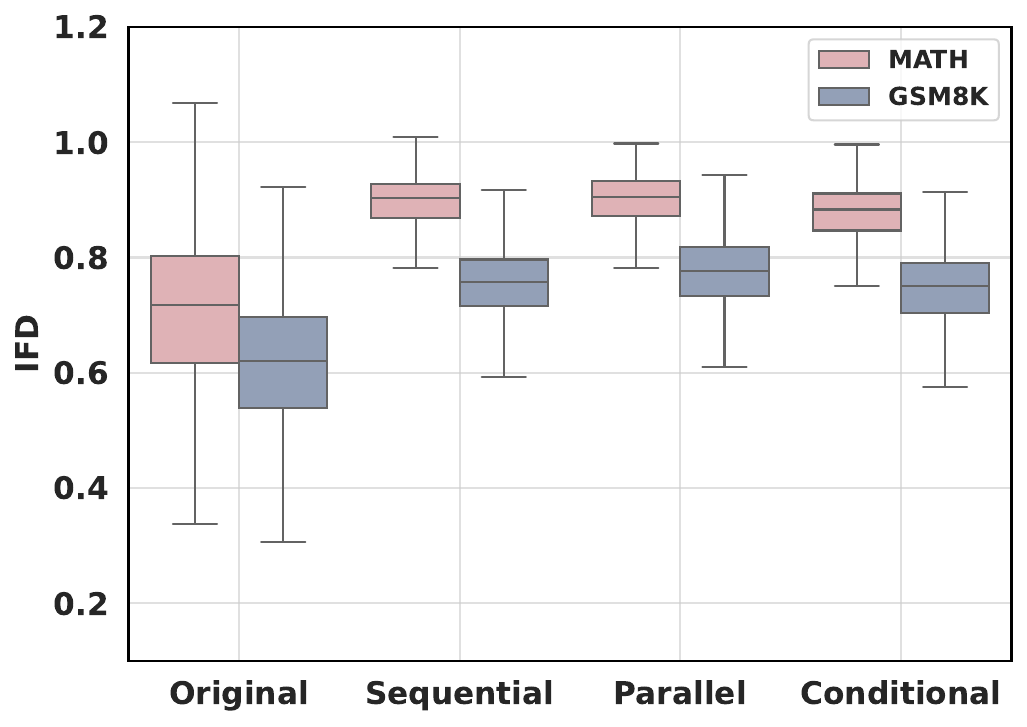}
    \label{fig:analysis_ifd}
  }  
    \subfigure{
      \includegraphics[width=0.305\linewidth]{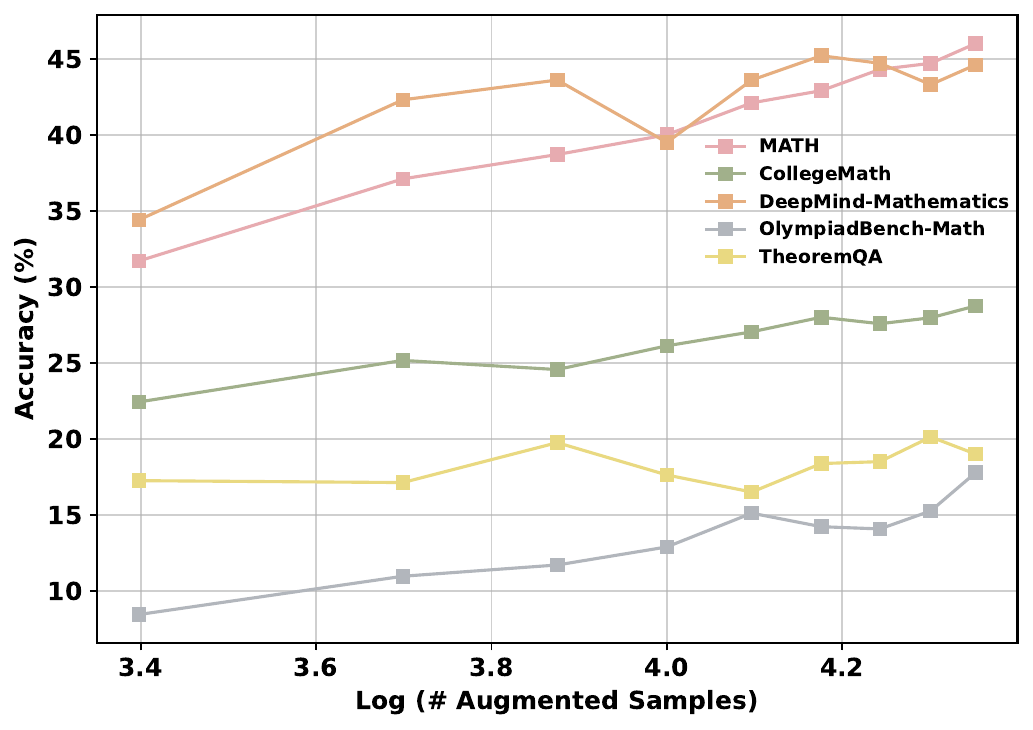}
      \label{fig:analysis_data_size_math}
    }
  \caption{
    \textbf{(a)}: Unconditional and conditional PPL for the original and fused data on GSM8K and MATH datasets.
    \textbf{(b)}: IFD for the original and fused data on GSM8K and MATH datasets.
    \textbf{(c)}: Performance scaling behavior of the \method~on different sizes of augmented data on Llama3-8B. 
    }
  \label{fig:combined_figure}
\end{figure*}

\begin{figure*}[t!]
  \centering
  \subfigure{
    \includegraphics[width=0.31\linewidth]{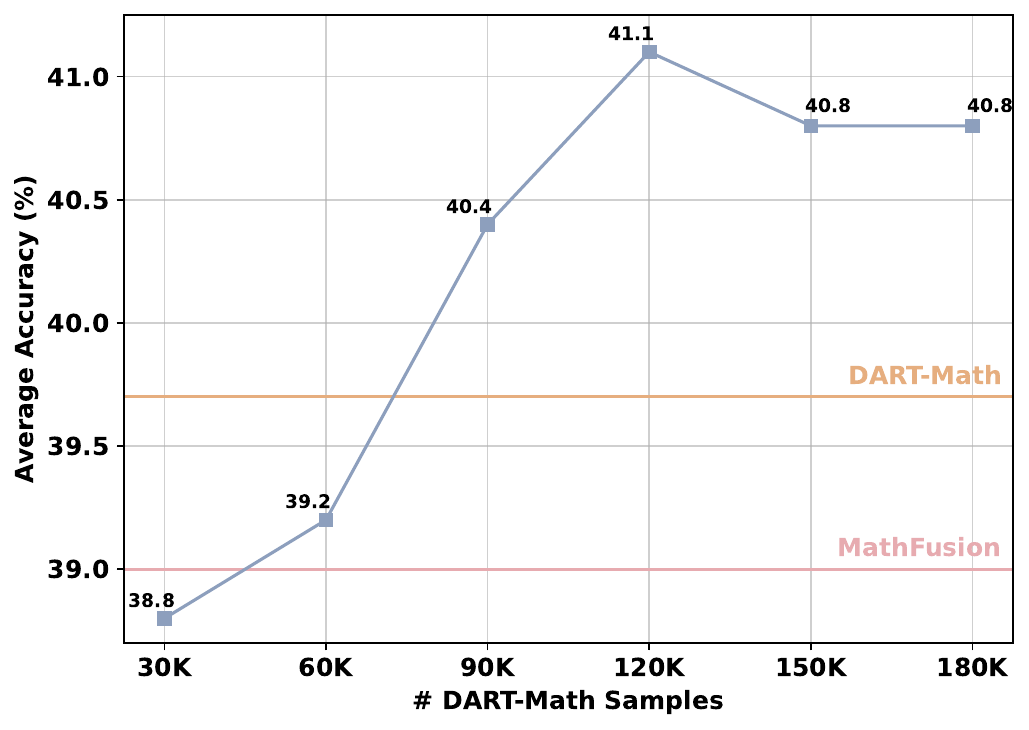}
    \label{fig:analysis_data_size_dart_math}
  }
  \subfigure{
    \includegraphics[width=0.31\linewidth]{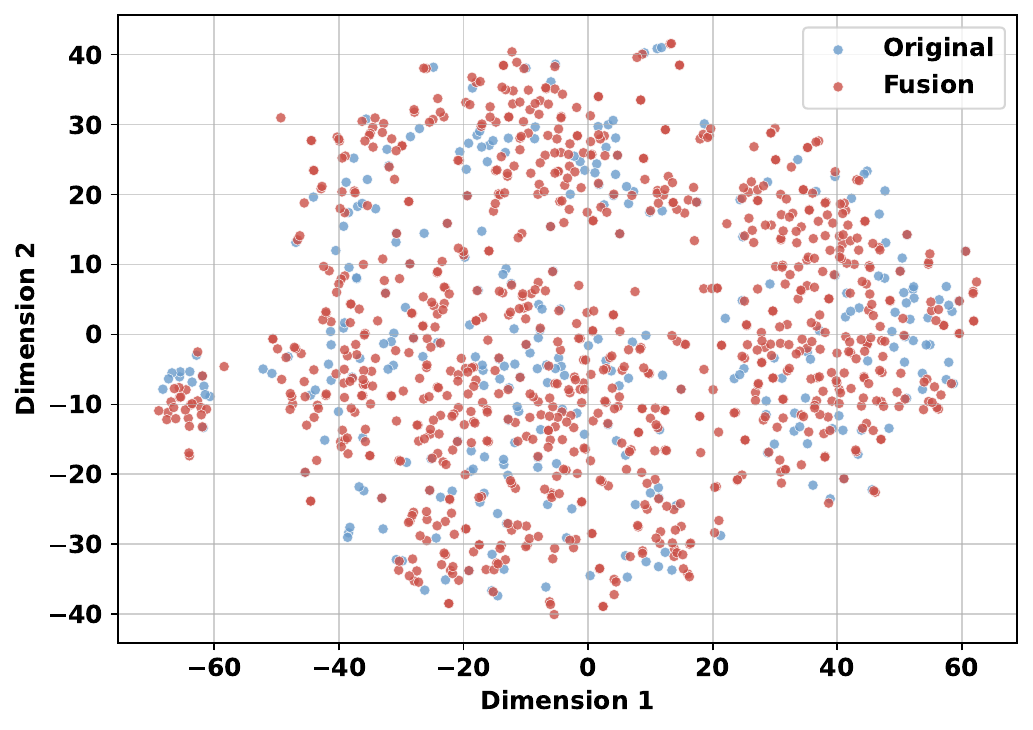}
    \label{fig:analysis_tsne_gsm8k}
  }
  \subfigure{
    \includegraphics[width=0.31\linewidth]{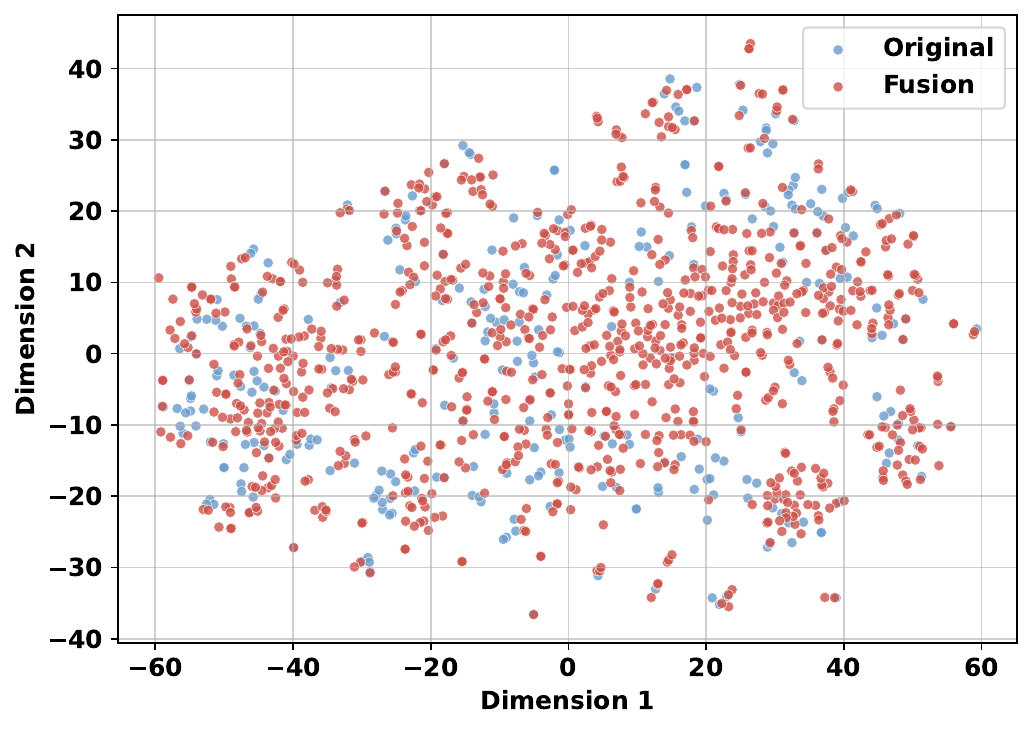}
    \label{fig:analysis_tsne_math}
  }
  \caption{
    \textbf{(a)}: Average performance of the Llama3-8B models fine-tuned on the combined dataset of \dataset~and DART-Math-Hard with different sizes of sampled data.
    \textbf{(b) and (c)}: Problem embedding visualization for GSM8K and MATH datasets via t-SNE.
    }
  \label{fig:combined_figure}
\end{figure*}

\subsection{Ablation Study}

\begin{table}[!t]
    \centering
    \resizebox{\linewidth}{!}{
    \begin{tabular}{c ccccc}
        \toprule
        Method & Sequential & Parallel & Conditional & MATH & GSM8K \\ \midrule
        Standard & \xmark & \xmark & \xmark & 17.5 & 65.4 \\
        \midrule
        \multirow{4}{*}{\method}
        & \xmark & \cmark & \cmark & 42.6 & 78.2  \\
        & \cmark & \xmark & \cmark & 43.0 & 76.9  \\
        & \cmark & \cmark & \xmark & 43.6 & 79.2 \\
        & \cmark & \cmark & \cmark & \textbf{45.6} & \textbf{79.9} \\
        \bottomrule
    \end{tabular}
    \vspace{-0.2cm}
    }
\caption{Effect of three fusion strategies on Llama3-8B.}
\label{tab:ablation_result_v1}
\vspace{-0.1cm}
\end{table}
We further conduct an ablation study to investigate the contribution of each fusion strategy to the overall performance of combined fusion.
The results over Llama3-8B on MATH and GSM8K are shown in Table~\ref{tab:ablation_result_v1}, from which we observe that each fusion strategy contributes to the overall performance, with \textit{conditional fusion} showing the least contribution, which aligns with Section~\ref{sec:main_results}.
We further ablate on choice of teacher model for solution generation in Appendix~\ref{sec:appendix_teacher_model}.

\section{Analysis}
We analyze the difficulty of the fused problem in Section~\ref{sec:problem_difficulty}, the relationship between augmented data size and performance in Section~\ref{sec:rel_aug_size_p}, the combination of \dataset{} with other datasets in Section~\ref{sec:comb_other_data}, and the diversity of fused problem in Section~\ref{sec:diversity_analy}.

\subsection{Difficulty Analysis}
\label{sec:problem_difficulty}
In this section, we explore why the three fusion strategies effectively enhance the model's performance.
To achieve this, we evaluate both the perplexity (PPL) and instruction following difficulty (IFD)~\citep{ifd} for the original and fused data. 
We use Mathstral-7B~\citep{mathstral}, a model built upon Mistral-7B~\citep{jiang2023mistral} and specifically fine-tuned for mathematical reasoning, to ensure our analysis relies on a model specifically designed for mathematical tasks.
Specifically, we denote the unconditioned PPL as $\text{PPL}(S)$, the conditioned PPL as $\text{PPL}(S\mid\! P)$, and $\text{IFD} = \text{PPL}(S\mid\! P) / \text{PPL}(S)$, where $P$ is the problem and $S$ is the solution. 
The results are shown in Figure~\ref{fig:analysis_ppl} and~\ref{fig:analysis_ifd}, from which we can see:
(1) The PPL of the solution of the fused problems is significantly lower than that of the original problems. As analyzed in~\citet{metamath}, this may be due to the easy-to-learn nature of the generated solutions.
(2) The IFD of the fused data is significantly higher than that of the original data, indicating that the fused data is more difficult to learn in the context of the problem.
(3) The IFD of the MATH datasets, both the original or fused version, are higher than that of the GSM8K, consistent with the fact that MATH is generally more difficult than GSM8K.

\subsection{Relationship between Augmented Data Size and Performance}
\label{sec:rel_aug_size_p}
We study the performance scaling behavior of the \method~on different sizes of augmented data on Llama3-8B.
We select MATH as the original training set and gradually increase the size of the augmented fusion data from 0 to 22.5K, with a step size of 2.5K.
The results on MATH and four OOD benchmarks are shown in Figure~\ref{fig:analysis_data_size_math}.
We observe that the performance of the \method~models exhibits an approximate logarithmic growth with respect to the amount of augmented data, which is consistent with the findings in~\citep{muggle_math}.
Additionally, the augmented fusion data from MATH dataset can also generalize better to the OOD benchmarks as the size of the augmented data increases.
In summary, the \method~shows consistent performance improvement with different sizes of augmented data.

\subsection{Combination with Other Datasets}
\label{sec:comb_other_data}
We further investigate the performance of \method~when combined with other data augmentation methods.
Specifically, we downsample 30K-180K data from DART-Math-Hard~\citep{dartmath}, which is the SOTA method for mathematical data augmentation with 590K data.
We combine the downsampled DART-Math-Hard with our \dataset~dataset and fine-tune Llama3-8B models on the combined dataset.
The results are presented in Figure~\ref{fig:analysis_data_size_dart_math}.
As the size of sampled data increases, the average performance of the models also increases, and reaches the peak when the size of the sampled data is 120K.
Notably, by only using 90K data sampled from DART-Math-Hard (\textit{i.e.}, 150K samples in total), the resulting model achieves better performance than both DART-Math and \method, yields SOTA average performance.
These results show the potential of combining \method~with other data augmentation methods to further enhance the model's performance.
We think that the enhancement arises from the complementary and orthogonal nature of the two methods: our \method~emphasizes fusing mathematical problems to generate more challenging and diverse problems, while DART-Math focuses on existing difficult problems and primarily generates additional solutions for them.

\subsection{Diversity Analysis}
\label{sec:diversity_analy}
To further investigate the effectiveness of the \method~in enhancing the data diversity, we visualize the problem embeddings of the GSM8K and MATH datasets generated by GPT-4o-mini using t-SNE~\citep{van2008tsne}.
The results are shown in Figure~\ref{fig:analysis_tsne_gsm8k} and~\ref{fig:analysis_tsne_math}.
We can observe that the \method~augmented problems are more evenly distributed in the embedding space, thereby enriching the diversity of the training examples and mitigating the risk of model overfitting.

\section{Conclusion}
In this paper, we focus on the fusion of mathematical problems. We propose a novel mathematical data augmentation method, \method, which comprises three distinct fusion strategies—\textit{sequential fusion}, \textit{parallel fusion}, and \textit{conditional fusion}—designed to synthesize augmented mathematical problems. 
Leveraging these fusion strategies, we construct the \dataset~dataset, which is subsequently employed to fine-tune LLMs. Extensive experiments on three base models and six benchmarks show that \method~exhibits robust performance in both the in-domain and out-of-domain benchmarks while maintaining high data efficiency.

\section*{Limitations}
We utilize GPT-4o-mini to generate fused problems and solutions, but the generated problems or solutions may still contain errors or ambiguities, which are hard to detect and verify.
The quality of the generated problems and solutions is limited by the capabilities of the teacher LLM.
Stronger teacher model, like DeepSeek-R1 and Qwen3, are underexplored.
We mainly explore the effectiveness of the three fusion strategies on problem pairs that are constructed by embedding similarity. The fusion of three or more problems and more effective ways to find similar problems, remain underexplored.
The released \dataset~dataset currently contains only 60K examples, and scaling to millions of examples remains underexplored.

\section*{Acknowledgements}
This work is supported by Beijing Outstanding Young Scientist Program NO. BJJWZYJH012019100020098 and by National Key R\&D Program of China (2022ZD0160201).
This work is also supported by the Public Computing Cloud, Renmin University of China and by fund for building world-class universities (disciplines) of Renmin University of China.
Qizhi Pei is supported by the Outstanding Innovative Talents Cultivation Funded Programs 2023 of Renmin University of China.
Qizhi Pei is an intern at Shanghai Artificial Intelligence Laboratory.

\bibliography{custom}

\begin{thebibliography}{55}
\providecommand{\natexlab}[1]{#1}

\bibitem[{Ahn et~al.(2024)Ahn, Verma, Lou, Liu, Zhang, and Yin}]{llm_math_survey}
Janice Ahn, Rishu Verma, Renze Lou, Di~Liu, Rui Zhang, and Wenpeng Yin. 2024.
\newblock Large language models for mathematical reasoning: Progresses and challenges.
\newblock In \emph{{EACL} (Student Research Workshop)}, pages 225--237. Association for Computational Linguistics.

\bibitem[{Bagherzadeh et~al.(2019)Bagherzadeh, Gurca, and Brunswicker}]{bagherzadeh2019problem}
Mehdi Bagherzadeh, Andrei Gurca, and Sabine Brunswicker. 2019.
\newblock Problem types and open innovation governance modes: A project-level empirical exploration.
\newblock \emph{IEEE Transactions on Engineering Management}, 69(2):287--301.

\bibitem[{Cao et~al.(2025)Cao, Zhou, Dai, Wang, and Zhang}]{mixup_survey1}
Chengtai Cao, Fan Zhou, Yurou Dai, Jianping Wang, and Kunpeng Zhang. 2025.
\newblock A survey of mix-based data augmentation: Taxonomy, methods, applications, and explainability.
\newblock \emph{{ACM} Comput. Surv.}, 57(2):37:1--37:38.

\bibitem[{Chen et~al.(2023)Chen, Yin, Ku, Lu, Wan, Ma, Xu, Wang, and Xia}]{chen2023theoremqa}
Wenhu Chen, Ming Yin, Max Ku, Pan Lu, Yixin Wan, Xueguang Ma, Jianyu Xu, Xinyi Wang, and Tony Xia. 2023.
\newblock \href {https://openreview.net/forum?id=Wom397PB55} {Theorem{QA}: A theorem-driven question answering dataset}.
\newblock In \emph{The 2023 Conference on Empirical Methods in Natural Language Processing}.

\bibitem[{Chu-Carroll et~al.(2024)Chu-Carroll, Beck, Burnham, Melville, Nachman, {\"O}zcan, and Ferrucci}]{chu2024beyond}
Jennifer Chu-Carroll, Andrew Beck, Greg Burnham, David~OS Melville, David Nachman, A~Erdem {\"O}zcan, and David Ferrucci. 2024.
\newblock Beyond llms: Advancing the landscape of complex reasoning.
\newblock \emph{arXiv preprint arXiv:2402.08064}.

\bibitem[{Cobbe et~al.(2021)Cobbe, Kosaraju, Bavarian, Chen, Jun, Kaiser, Plappert, Tworek, Hilton, Nakano et~al.}]{cobbe2021gsm8k}
Karl Cobbe, Vineet Kosaraju, Mohammad Bavarian, Mark Chen, Heewoo Jun, Lukasz Kaiser, Matthias Plappert, Jerry Tworek, Jacob Hilton, Reiichiro Nakano, et~al. 2021.
\newblock Training verifiers to solve math word problems.
\newblock \emph{arXiv preprint arXiv:2110.14168}.

\bibitem[{Ding et~al.(2024)Ding, Shi, Liang, Li, Zhu, and Zhang}]{scale_quest}
Yuyang Ding, Xinyu Shi, Xiaobo Liang, Juntao Li, Qiaoming Zhu, and Min Zhang. 2024.
\newblock Unleashing reasoning capability of llms via scalable question synthesis from scratch.
\newblock \emph{arXiv preprint arXiv:2410.18693}.

\bibitem[{Dubey et~al.(2024)Dubey, Jauhri, Pandey, Kadian, Al-Dahle, Letman, Mathur, Schelten, Yang, Fan et~al.}]{llama3}
Abhimanyu Dubey, Abhinav Jauhri, Abhinav Pandey, Abhishek Kadian, Ahmad Al-Dahle, Aiesha Letman, Akhil Mathur, Alan Schelten, Amy Yang, Angela Fan, et~al. 2024.
\newblock The llama 3 herd of models.
\newblock \emph{arXiv preprint arXiv:2407.21783}.

\bibitem[{Guan et~al.(2025)Guan, Zhang, Liu, Shang, Sun, Zhu, Yang, and Yang}]{rstar_math}
Xinyu Guan, Li~Lyna Zhang, Yifei Liu, Ning Shang, Youran Sun, Yi~Zhu, Fan Yang, and Mao Yang. 2025.
\newblock rstar-math: Small llms can master math reasoning with self-evolved deep thinking.
\newblock \emph{arXiv preprint arXiv:2501.04519}.

\bibitem[{Guo et~al.(2020)Guo, Kim, and Rush}]{guo2020sequence}
Demi Guo, Yoon Kim, and Alexander~M. Rush. 2020.
\newblock Sequence-level mixed sample data augmentation.
\newblock In \emph{{EMNLP} {(1)}}, pages 5547--5552. Association for Computational Linguistics.

\bibitem[{Guo et~al.(2019)Guo, Mao, and Zhang}]{guo2019augmenting}
Hongyu Guo, Yongyi Mao, and Richong Zhang. 2019.
\newblock Augmenting data with mixup for sentence classification: An empirical study.
\newblock \emph{arXiv preprint arXiv:1905.08941}.

\bibitem[{He et~al.(2024)He, Luo, Bai, Hu, Thai, Shen, Hu, Han, Huang, Zhang et~al.}]{he2024olympiadbench}
Chaoqun He, Renjie Luo, Yuzhuo Bai, Shengding Hu, Zhen~Leng Thai, Junhao Shen, Jinyi Hu, Xu~Han, Yujie Huang, Yuxiang Zhang, et~al. 2024.
\newblock Olympiadbench: A challenging benchmark for promoting agi with olympiad-level bilingual multimodal scientific problems.
\newblock \emph{arXiv preprint arXiv:2402.14008}.

\bibitem[{Hendrycks et~al.(2021)Hendrycks, Burns, Kadavath, Arora, Basart, Tang, Song, and Steinhardt}]{hendrycks2021math}
Dan Hendrycks, Collin Burns, Saurav Kadavath, Akul Arora, Steven Basart, Eric Tang, Dawn Song, and Jacob Steinhardt. 2021.
\newblock \href {https://datasets-benchmarks-proceedings.neurips.cc/paper_files/paper/2021/file/be83ab3ecd0db773eb2dc1b0a17836a1-Paper-round2.pdf} {Measuring mathematical problem solving with the math dataset}.
\newblock In \emph{Proceedings of the Neural Information Processing Systems Track on Datasets and Benchmarks}, volume~1.

\bibitem[{Huang and Chang(2023)}]{llm_reasoning_survey}
Jie Huang and Kevin~Chen{-}Chuan Chang. 2023.
\newblock Towards reasoning in large language models: {A} survey.
\newblock In \emph{{ACL} (Findings)}, pages 1049--1065. Association for Computational Linguistics.

\bibitem[{Huang et~al.(2024)Huang, Liu, Gong, Gou, Shen, Duan, and Chen}]{kpmath}
Yiming Huang, Xiao Liu, Yeyun Gong, Zhibin Gou, Yelong Shen, Nan Duan, and Weizhu Chen. 2024.
\newblock Key-point-driven data synthesis with its enhancement on mathematical reasoning.
\newblock \emph{arXiv preprint arXiv:2403.02333}.

\bibitem[{Jiang et~al.(2023)Jiang, Sablayrolles, Mensch, Bamford, Chaplot, de~las Casas, Bressand, Lengyel, Lample, Saulnier, Lavaud, Lachaux, Stock, Scao, Lavril, Wang, Lacroix, and Sayed}]{jiang2023mistral}
Albert~Q. Jiang, Alexandre Sablayrolles, Arthur Mensch, Chris Bamford, Devendra~Singh Chaplot, Diego de~las Casas, Florian Bressand, Gianna Lengyel, Guillaume Lample, Lucile Saulnier, Lélio~Renard Lavaud, Marie-Anne Lachaux, Pierre Stock, Teven~Le Scao, Thibaut Lavril, Thomas Wang, Timothée Lacroix, and William~El Sayed. 2023.
\newblock \href {https://arxiv.org/abs/2310.06825} {Mistral 7b}.
\newblock \emph{Preprint}, arXiv:2310.06825.

\bibitem[{Jin et~al.(2024)Jin, Zhu, Li, Wang, Liu, Yu, Qin, and Li}]{mixup_survey2}
Xin Jin, Hongyu Zhu, Siyuan Li, Zedong Wang, Zicheng Liu, Chang Yu, Huafeng Qin, and Stan~Z Li. 2024.
\newblock A survey on mixup augmentations and beyond.
\newblock \emph{arXiv preprint arXiv:2409.05202}.

\bibitem[{Kang et~al.(2024)Kang, Li, Chen, Kazemi, Sun, Chen, Li, He, He, Wen et~al.}]{mindstar}
Jikun Kang, Xin~Zhe Li, Xi~Chen, Amirreza Kazemi, Qianyi Sun, Boxing Chen, Dong Li, Xu~He, Quan He, Feng Wen, et~al. 2024.
\newblock Mindstar: Enhancing math reasoning in pre-trained llms at inference time.
\newblock \emph{arXiv preprint arXiv:2405.16265}.

\bibitem[{Kaur et~al.()Kaur, Park, Goyal, and Arora}]{instruct_skillmix}
Simran Kaur, Simon Park, Anirudh Goyal, and Sanjeev Arora.
\newblock Instruct-skillmix: A powerful pipeline for llm instruction tuning.
\newblock In \emph{NeurIPS 2024 Workshop on Fine-Tuning in Modern Machine Learning: Principles and Scalability}.

\bibitem[{Komarudin et~al.(2021)Komarudin, Suherman, and Anggraini}]{komarudin2021analysis}
Komarudin Komarudin, Suherman Suherman, and Anita Anggraini. 2021.
\newblock Analysis of mathematical concept understanding capabilities: The impact of makerspae stem learning approach models and student learning activities.
\newblock \emph{Journal of Innovation in Educational and Cultural Research}, 2(1):35--43.

\bibitem[{Li et~al.(2024{\natexlab{a}})Li, Wang, Hu, Wei, Zheng, Hu, Zhang, and Peng}]{xwin_math}
Chen Li, Weiqi Wang, Jingcheng Hu, Yixuan Wei, Nanning Zheng, Han Hu, Zheng Zhang, and Houwen Peng. 2024{\natexlab{a}}.
\newblock Common 7b language models already possess strong math capabilities.
\newblock \emph{arXiv preprint arXiv:2403.04706}.

\bibitem[{Li et~al.(2024{\natexlab{b}})Li, Yuan, Yuan, Dong, Lu, Wu, Tan, Wang, and Zhou}]{muggle_math}
Chengpeng Li, Zheng Yuan, Hongyi Yuan, Guanting Dong, Keming Lu, Jiancan Wu, Chuanqi Tan, Xiang Wang, and Chang Zhou. 2024{\natexlab{b}}.
\newblock Mugglemath: Assessing the impact of query and response augmentation on math reasoning.
\newblock In \emph{{ACL} {(1)}}, pages 10230--10258. Association for Computational Linguistics.

\bibitem[{Li et~al.(2024{\natexlab{c}})Li, Chen, Wang, Zhao, Liang, Hou, Liu, and Zhou}]{mosaic}
Ming Li, Pei Chen, Chenguang Wang, Hongyu Zhao, Yijun Liang, Yupeng Hou, Fuxiao Liu, and Tianyi Zhou. 2024{\natexlab{c}}.
\newblock Mosaic it: Enhancing instruction tuning with data mosaics.
\newblock \emph{arXiv preprint arXiv:2405.13326}.

\bibitem[{Li et~al.(2024{\natexlab{d}})Li, Zhang, Li, Chen, Chen, Cheng, Wang, Zhou, and Xiao}]{ifd}
Ming Li, Yong Zhang, Zhitao Li, Jiuhai Chen, Lichang Chen, Ning Cheng, Jianzong Wang, Tianyi Zhou, and Jing Xiao. 2024{\natexlab{d}}.
\newblock From quantity to quality: Boosting {LLM} performance with self-guided data selection for instruction tuning.
\newblock In \emph{{NAACL-HLT}}, pages 7602--7635. Association for Computational Linguistics.

\bibitem[{Lin et~al.(2025)Lin, Pan, Li, Pei, Gao, Cai, He, and Wu}]{lin2025metaladder}
Honglin Lin, Zhuoshi Pan, Yu~Li, Qizhi Pei, Xin Gao, Mengzhang Cai, Conghui He, and Lijun Wu. 2025.
\newblock Metaladder: Ascending mathematical solution quality via analogical-problem reasoning transfer.
\newblock \emph{arXiv preprint arXiv:2503.14891}.

\bibitem[{Liu et~al.(2024)Liu, Zhang, Luo, and Yao}]{liu2024mmiqc}
Haoxiong Liu, Yifan Zhang, Yifan Luo, and Andrew Chi-Chih Yao. 2024.
\newblock \href {https://arxiv.org/abs/2401.09003} {Augmenting math word problems via iterative question composing}.
\newblock \emph{Preprint}, arXiv:2401.09003.

\bibitem[{Lu et~al.(2024)Lu, Zhou, Wang, Ren, Shi, Pan, Zhan, and Li}]{step_control_dpo}
Zimu Lu, Aojun Zhou, Ke~Wang, Houxing Ren, Weikang Shi, Junting Pan, Mingjie Zhan, and Hongsheng Li. 2024.
\newblock Step-controlled dpo: Leveraging stepwise error for enhanced mathematical reasoning.
\newblock \emph{arXiv preprint arXiv:2407.00782}.

\bibitem[{Luo et~al.(2023)Luo, Sun, Xu, Zhao, Lou, Tao, Geng, Lin, Chen, and Zhang}]{wizardmath}
Haipeng Luo, Qingfeng Sun, Can Xu, Pu~Zhao, Jianguang Lou, Chongyang Tao, Xiubo Geng, Qingwei Lin, Shifeng Chen, and Dongmei Zhang. 2023.
\newblock Wizardmath: Empowering mathematical reasoning for large language models via reinforced evol-instruct.
\newblock \emph{arXiv preprint arXiv:2308.09583}.

\bibitem[{Mitra et~al.(2024)Mitra, Khanpour, Rosset, and Awadallah}]{mitra2024orcamath}
Arindam Mitra, Hamed Khanpour, Corby Rosset, and Ahmed Awadallah. 2024.
\newblock Orca-math: Unlocking the potential of slms in grade school math.
\newblock \emph{arXiv preprint arXiv:2402.14830}.

\bibitem[{OpenAI et~al.(2023)OpenAI, Adler, Agarwal, Ahmad, Akkaya, Aleman, Almeida, Altenschmidt, Altman, Anadkat et~al.}]{openai2023gpt4}
Josh OpenAI, Achiam, Steven Adler, Sandhini Agarwal, Lama Ahmad, Ilge Akkaya, Florencia~Leoni Aleman, Diogo Almeida, Janko Altenschmidt, Sam Altman, Shyamal Anadkat, et~al. 2023.
\newblock Gpt-4 technical report.
\newblock \emph{arXiv preprint arXiv:2303.08774}.

\bibitem[{Pan et~al.(2025)Pan, Li, Lin, Pei, Tang, Wu, Ming, Zhao, He, and Wu}]{pan2025lemma}
Zhuoshi Pan, Yu~Li, Honglin Lin, Qizhi Pei, Zinan Tang, Wei Wu, Chenlin Ming, H~Vicky Zhao, Conghui He, and Lijun Wu. 2025.
\newblock Lemma: Learning from errors for mathematical advancement in llms.
\newblock \emph{arXiv preprint arXiv:2503.17439}.

\bibitem[{Prabawa et~al.(2023)Prabawa, Rosjanuardi, and Nurlaelah}]{prabawa2023problem}
Harsa~Wara Prabawa, Rizky Rosjanuardi, and Elah Nurlaelah. 2023.
\newblock Problem decomposition skills, mathematical maturity, and their relation to mathematics problem-solving in a computer science learning class.
\newblock \emph{Jurnal Kependidikan: Jurnal Hasil Penelitian dan Kajian Kepustakaan di Bidang Pendidikan, Pengajaran dan Pembelajaran}, 9(3):946--958.

\bibitem[{Saxton et~al.(2019)Saxton, Grefenstette, Hill, and Kohli}]{saxton2018dmmath}
David Saxton, Edward Grefenstette, Felix Hill, and Pushmeet Kohli. 2019.
\newblock \href {https://openreview.net/forum?id=H1gR5iR5FX} {Analysing mathematical reasoning abilities of neural models}.
\newblock In \emph{International Conference on Learning Representations}.

\bibitem[{Setlur et~al.(2024)Setlur, Garg, Geng, Garg, Smith, and Kumar}]{setlur2024rl}
Amrith Setlur, Saurabh Garg, Xinyang Geng, Naman Garg, Virginia Smith, and Aviral Kumar. 2024.
\newblock Rl on incorrect synthetic data scales the efficiency of llm math reasoning by eight-fold.
\newblock \emph{arXiv preprint arXiv:2406.14532}.

\bibitem[{Shao et~al.(2024)Shao, Wang, Zhu, Xu, Song, Bi, Zhang, Zhang, Li, Wu et~al.}]{deepseekmath}
Zhihong Shao, Peiyi Wang, Qihao Zhu, Runxin Xu, Junxiao Song, Xiao Bi, Haowei Zhang, Mingchuan Zhang, YK~Li, Y~Wu, et~al. 2024.
\newblock Deepseekmath: Pushing the limits of mathematical reasoning in open language models.
\newblock \emph{arXiv preprint arXiv:2402.03300}.

\bibitem[{Srivatsa and Kochmar(2024)}]{srivatsa2024makes}
KV~Aditya Srivatsa and Ekaterina Kochmar. 2024.
\newblock What makes math word problems challenging for llms?
\newblock In \emph{{NAACL-HLT} (Findings)}, pages 1138--1148. Association for Computational Linguistics.

\bibitem[{Tang et~al.(2024)Tang, Zhang, Wang, and Wei}]{mathscale}
Zhengyang Tang, Xingxing Zhang, Benyou Wang, and Furu Wei. 2024.
\newblock Mathscale: Scaling instruction tuning for mathematical reasoning.
\newblock In \emph{{ICML}}. OpenReview.net.

\bibitem[{team(2024)}]{mathstral}
Mistral~AI team. 2024.
\newblock \href {https://mistral.ai/en/news/mathstral} {Learning to reason with llms}.

\bibitem[{Thulasidasan et~al.(2019)Thulasidasan, Chennupati, Bilmes, Bhattacharya, and Michalak}]{mixup_train}
Sunil Thulasidasan, Gopinath Chennupati, Jeff~A. Bilmes, Tanmoy Bhattacharya, and Sarah Michalak. 2019.
\newblock On mixup training: Improved calibration and predictive uncertainty for deep neural networks.
\newblock In \emph{NeurIPS}, pages 13888--13899.

\bibitem[{Tong et~al.(2024)Tong, Zhang, Wang, Wu, and He}]{dartmath}
Yuxuan Tong, Xiwen Zhang, Rui Wang, Ruidong Wu, and Junxian He. 2024.
\newblock Dart-math: Difficulty-aware rejection tuning for mathematical problem-solving.
\newblock In \emph{NeurIPS}.

\bibitem[{Toshniwal et~al.()Toshniwal, Du, Moshkov, Kisacanin, Ayrapetyan, and Gitman}]{toshniwalopenmathinstruct-2}
Shubham Toshniwal, Wei Du, Ivan Moshkov, Branislav Kisacanin, Alexan Ayrapetyan, and Igor Gitman.
\newblock Openmathinstruct-2: Accelerating ai for math with massive open-source instruction data.
\newblock In \emph{The Thirteenth International Conference on Learning Representations}.

\bibitem[{Van~der Maaten and Hinton(2008)}]{van2008tsne}
Laurens Van~der Maaten and Geoffrey Hinton. 2008.
\newblock Visualizing data using t-sne.
\newblock \emph{Journal of machine learning research}, 9(11).

\bibitem[{Wang et~al.(2024{\natexlab{a}})Wang, Ren, Zhou, Lu, Luo, Shi, Zhang, Song, Zhan, and Li}]{mathcoder}
Ke~Wang, Houxing Ren, Aojun Zhou, Zimu Lu, Sichun Luo, Weikang Shi, Renrui Zhang, Linqi Song, Mingjie Zhan, and Hongsheng Li. 2024{\natexlab{a}}.
\newblock Mathcoder: Seamless code integration in llms for enhanced mathematical reasoning.
\newblock In \emph{{ICLR}}. OpenReview.net.

\bibitem[{Wang et~al.(2024{\natexlab{b}})Wang, Liu, Liang, Li, Huang, Zhang, Shen, Guan, Wang, Feng et~al.}]{wang2024langgpt}
Ming Wang, Yuanzhong Liu, Xiaoyu Liang, Songlian Li, Yijie Huang, Xiaoming Zhang, Sijia Shen, Chaofeng Guan, Daling Wang, Shi Feng, et~al. 2024{\natexlab{b}}.
\newblock Langgpt: Rethinking structured reusable prompt design framework for llms from the programming language.
\newblock \emph{arXiv preprint arXiv:2402.16929}.

\bibitem[{Wei et~al.(2022)Wei, Wang, Schuurmans, Bosma, Ichter, Xia, Chi, Le, and Zhou}]{cot_reasoning}
Jason Wei, Xuezhi Wang, Dale Schuurmans, Maarten Bosma, Brian Ichter, Fei Xia, Ed~H. Chi, Quoc~V. Le, and Denny Zhou. 2022.
\newblock Chain-of-thought prompting elicits reasoning in large language models.
\newblock In \emph{NeurIPS}.

\bibitem[{Wu et~al.(2021)Wu, Xia, Zhu, Wu, Xie, Fan, and Qin}]{wu2021mixseq}
Xueqing Wu, Yingce Xia, Jinhua Zhu, Lijun Wu, Shufang Xie, Yang Fan, and Tao Qin. 2021.
\newblock mixseq: {A} simple data augmentation methodfor neural machine translation.
\newblock In \emph{{IWSLT}}, pages 192--197. Association for Computational Linguistics.

\bibitem[{Xi et~al.(2024)Xi, Yang, Huang, Tang, Li, Ding, He, Hong, Do, Zhan et~al.}]{mathcritique}
Zhiheng Xi, Dingwen Yang, Jixuan Huang, Jiafu Tang, Guanyu Li, Yiwen Ding, Wei He, Boyang Hong, Shihan Do, Wenyu Zhan, et~al. 2024.
\newblock Enhancing llm reasoning via critique models with test-time and training-time supervision.
\newblock \emph{arXiv preprint arXiv:2411.16579}.

\bibitem[{Yang et~al.(2024)Yang, Zhang, Hui, Gao, Yu, Li, Liu, Tu, Zhou, Lin et~al.}]{qwen25_math}
An~Yang, Beichen Zhang, Binyuan Hui, Bofei Gao, Bowen Yu, Chengpeng Li, Dayiheng Liu, Jianhong Tu, Jingren Zhou, Junyang Lin, et~al. 2024.
\newblock Qwen2. 5-math technical report: Toward mathematical expert model via self-improvement.
\newblock \emph{arXiv preprint arXiv:2409.12122}.

\bibitem[{Ying et~al.(2024)Ying, Zhang, Li, Zhou, Shao, Fei, Ma, Hong, Liu, Wang et~al.}]{internlm_math}
Huaiyuan Ying, Shuo Zhang, Linyang Li, Zhejian Zhou, Yunfan Shao, Zhaoye Fei, Yichuan Ma, Jiawei Hong, Kuikun Liu, Ziyi Wang, et~al. 2024.
\newblock Internlm-math: Open math large language models toward verifiable reasoning.
\newblock \emph{arXiv preprint arXiv:2402.06332}.

\bibitem[{Yu et~al.(2024)Yu, Jiang, Shi, YU, Liu, Zhang, Kwok, Li, Weller, and Liu}]{metamath}
Longhui Yu, Weisen Jiang, Han Shi, Jincheng YU, Zhengying Liu, Yu~Zhang, James Kwok, Zhenguo Li, Adrian Weller, and Weiyang Liu. 2024.
\newblock \href {https://openreview.net/forum?id=N8N0hgNDRt} {Metamath: Bootstrap your own mathematical questions for large language models}.
\newblock In \emph{The Twelfth International Conference on Learning Representations}.

\bibitem[{Yuan et~al.(2023)Yuan, Yuan, Li, Dong, Tan, and Zhou}]{yuan2023rft}
Zheng Yuan, Hongyi Yuan, Chengpeng Li, Guanting Dong, Chuanqi Tan, and Chang Zhou. 2023.
\newblock Scaling relationship on learning mathematical reasoning with large language models.
\newblock \emph{arXiv preprint arXiv:2308.01825}.

\bibitem[{Zhang et~al.(2018)Zhang, Ciss{\'{e}}, Dauphin, and Lopez{-}Paz}]{mixup}
Hongyi Zhang, Moustapha Ciss{\'{e}}, Yann~N. Dauphin, and David Lopez{-}Paz. 2018.
\newblock mixup: Beyond empirical risk minimization.
\newblock In \emph{{ICLR} (Poster)}. OpenReview.net.

\bibitem[{Zhang et~al.(2020)Zhang, Yu, and Zhang}]{zhang2020seqmix}
Rongzhi Zhang, Yue Yu, and Chao Zhang. 2020.
\newblock Seqmix: Augmenting active sequence labeling via sequence mixup.
\newblock In \emph{{EMNLP} {(1)}}, pages 8566--8579. Association for Computational Linguistics.

\bibitem[{Zhang et~al.(2024)Zhang, Ge, Liang, Yu, Yu, Jia, Yu, and Jiang}]{refaug}
Zhihan Zhang, Tao Ge, Zhenwen Liang, Wenhao Yu, Dian Yu, Mengzhao Jia, Dong Yu, and Meng Jiang. 2024.
\newblock Learn beyond the answer: Training language models with reflection for mathematical reasoning.
\newblock In \emph{{EMNLP}}, pages 14720--14738. Association for Computational Linguistics.

\bibitem[{Zheng et~al.(2024)Zheng, Zhang, Zhang, Ye, Luo, Feng, and Ma}]{zheng2024llamafactory}
Yaowei Zheng, Richong Zhang, Junhao Zhang, Yanhan Ye, Zheyan Luo, Zhangchi Feng, and Yongqiang Ma. 2024.
\newblock \href {http://arxiv.org/abs/2403.13372} {Llamafactory: Unified efficient fine-tuning of 100+ language models}.
\newblock In \emph{Proceedings of the 62nd Annual Meeting of the Association for Computational Linguistics (Volume 3: System Demonstrations)}, Bangkok, Thailand. Association for Computational Linguistics.

\end{thebibliography}

\appendix
\clearpage
\section{Prompts}
\label{sec:appendix_prompts}
We show the prompts used for \textit{Sequential Fusion} in Prompt~\ref{trainprompt:seq_fusion}, \textit{Parallel Fusion} in Prompt~\ref{trainprompt:para_fusion}, and \textit{Conditional Fusion} in Prompt~\ref{trainprompt:cond_fusion}.
We also provide the problem evaluation prompts in Prompt~\ref{trainprompt:problem_eval}, which is partially derived from WizardMath~\citep{wizardmath}.
We use LangGPT~\citep{wang2024langgpt} to format prompts in Markdown and polish them.

\section{General Settings}
\subsection{Data Synthesis}
\label{sec:appendix_synthesis_setup}
We synthesize the augmented data, both the fusion process and the generation of the corresponding solutions, using GPT-4o-mini(\textit{gpt-4o-mini-2024-07-18})~\citep{openai2023gpt4}.
We set the temperature to 0.7 and the maximum length of generation to 4096.
The statistics of the generated data, as well as the base GSM8K~\citep{cobbe2021gsm8k} and MATH~\citep{hendrycks2021math} datasets, are shown in Table~\ref{tab:mathfusionqa_stat}.
\begin{table}[h]
  \centering
  \resizebox{\linewidth}{!}{
    \begin{tabular}{l ccc}
      \toprule
      Dataset & GSM8K & MATH & Total \\
      \midrule
      Standard & 7.5K & 7.5K & 15K \\
      \midrule
      \dataset~(\textit{Sequential}) & 15K & 15K & 30K \\
      \dataset~(\textit{Parallel}) & 15K & 15K & 30K \\
      \dataset~(\textit{Conditional}) & 15K & 15K & 30K \\
      \dataset & 30K & 30K & 60K \\
      \bottomrule
  \end{tabular}}
  \caption{
    Statistics of the \dataset~dataset and the original datasets GSM8K and MATH.
  }
  \label{tab:mathfusionqa_stat}
\end{table}

\subsection{Training}
\label{sec:appendix_training_setup}
We use LLaMA-Factory~\citep{zheng2024llamafactory} to fine-tune the models.
All models, including our own reproductions of baselines, are fine-tuned for 3 epochs with a batch size of 128 on 8xNVIDIA A100 GPU.
The peak learning rate is 5e-6 with a linear warm-up for the first 3\% of the training steps, followed by cosine decay.
The maximum sequence length is set to 4096.

In Table~\ref{tab:main_result}, we reproduce the results of the baselines with 60K data.
For MetaMath~\citep{metamath}, MMIQC~\citep{liu2024mmiqc}, and DART-Math~\citep{dartmath}, we directly downsample 60K data from the original datasets randomly.
For RefAug~\citep{refaug}, the original training set only contains 30K data, with 15K from GSM8K and MATH, and 15K from the augmented reflection data.
To upsample the RefAug dataset to 60K, we re-generate the reflection data two times using GPT-4o-mini with the original prompts~\citep{refaug}, thus obtaining an additional 30K data and forming the 60K dataset.

\subsection{Evaluation}
\label{sec:appendix_evaluation_setup}
We compare \method~models with baselines on the following six benchmarks:
\begin{itemize}
  \item \textbf{GSM8K}~\citep{cobbe2021gsm8k} dataset includes 8,792 high-quality grade school math word problems, with 7,473 for training and 1,319 for testing. Each problem in GSM8K requires between 2 and 8 steps to solve.
  \item \textbf{MATH}~\citep{hendrycks2021math} dataset is composed of 12,500 problems from high school math competitions, with 7,500 for training and 5,000 for testing. Problems in MATH are categorized into 7 types (Prealgebra, Intermediate Algebra, Algebra, Precalculus, Geometry, Counting \& Probability, and Number Theory) and 5 difficulty levels.
  \item \textbf{CollegeMath}~\citep{mathscale} test set contains 2,818 college-level problems, which are curated from 9 college-level mathematics textbooks, covering 7 key mathematical disciplines: Algebra, Precalculus, Calculus, VectorCalculus, Probability, LinearAlgebra, and Differential Equations.
  \item \textbf{DeepMind-Mathematics}~\citep{saxton2018dmmath} test set consists of 1,000 problems covering a wide range of mathematical reasoning tasks spanning algebra, arithmetic, calculus, and probability designed to evaluate the mathematical reasoning abilities of models.
  \item \textbf{OlympiadBench-Math}~\citep{he2024olympiadbench} benchmark including 675 Olympiad-level mathematical problems, and we only use the text-only English subset of Olympiad-Bench.
  \item \textbf{TheoremQA}~\citep{chen2023theoremqa} is a novel theorem-driven question-answering benchmark containing 800 problems based on 350 theorems. It is designed to evaluate LLM's ability to apply domain-specific theorems across fields such as Mathematics, Physics, Electrical Engineering, Computer Science, and Finance.
\end{itemize}

\subsection{Templates}
\label{sec:appendix_prompt_template}
For most of the results from our own runs, we use the template
\textit{"Question: \{problem\}\textbackslash nAnswer:"} for training, and
\textit{"Question: \{problem\}\textbackslash nAnswer: Let's think step by step."} for evaluation.
There are two exceptions:
(1) For reproduced DART-Math~\citep{dartmath}, we use its default Alpaca template: \textit{"Below is an instruction that describes a task. Write a response that appropriately completes the request.\textbackslash n\textbackslash n\#\#\#Instruction:\textbackslash n\{problem\}\textbackslash n\textbackslash n\#\#\# Response:\textbackslash n"}.
(2) For evaluation on the DeepMind Mathematics benchmark for models fine-tuned from Llama3-8B, we find the Alpaca template yields consistently better performance than the template above. Therefore we use the Alpaca template for all the Llama3-8B evaluation on this dataset.

\section{Analysis of Fused Problems}
The embedding search naturally ensures a high degree of contextual similarity. 
In the following sections, we analyze the fused problems in terms of problem types and errors.

\subsection{Fused Probelm Types}
\begin{figure*}
  \centering
  \includegraphics[width=\linewidth]{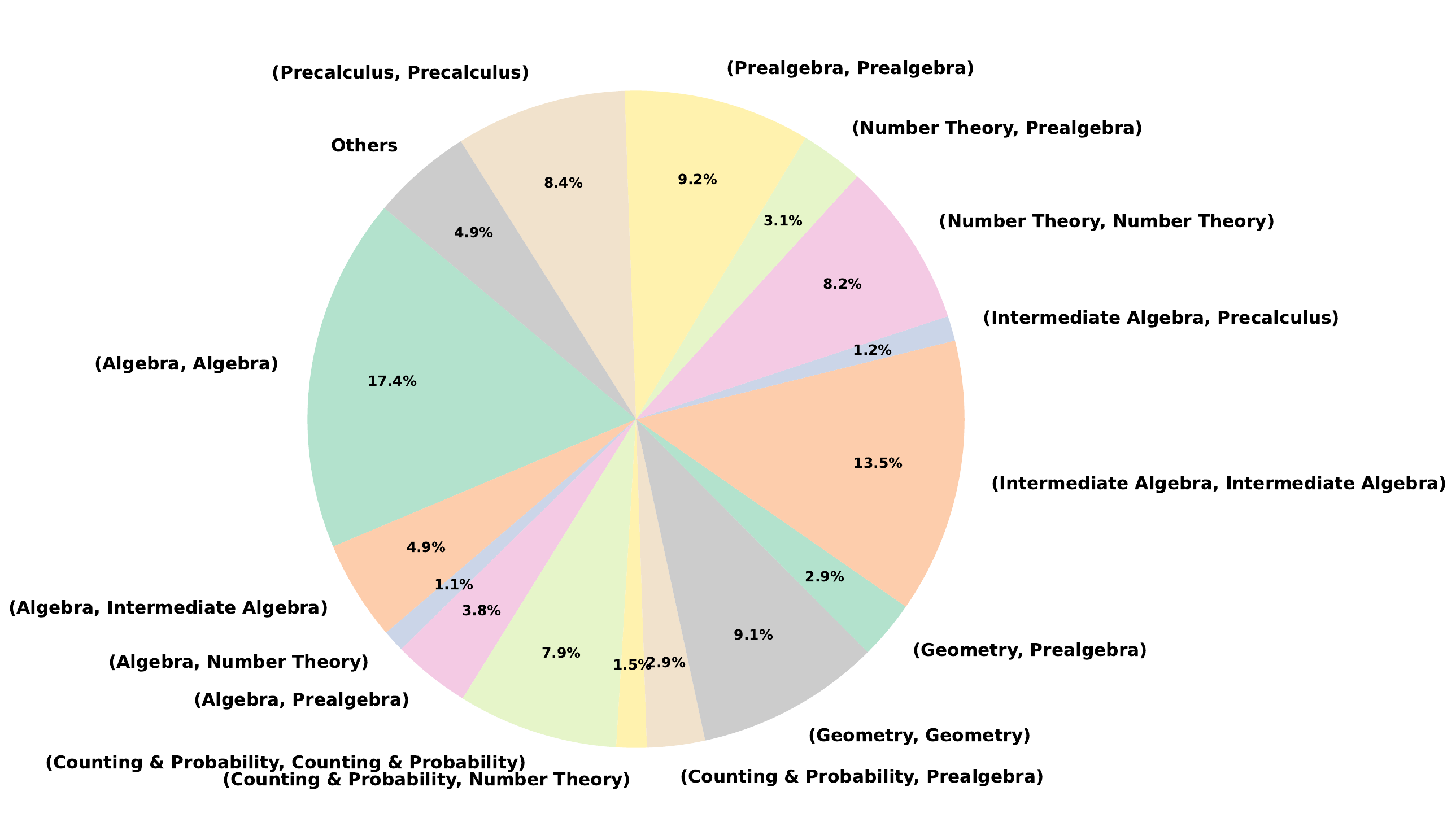}
  \caption{Distribution of combination types of problems in MATH dataset.}
  \label{fig:comb_math_type}
\end{figure*}

Regarding problem types, in the GSM8K~\citep{cobbe2021gsm8k} dataset, all problems are simple algebra questions. 
For the MATH dataset, we find that 83\% of the problem pairs belong to the same category, further validating the feasibility of the embedding search.
We plot the distribution of combination types of problems in MATH in Figure~\ref{fig:comb_math_type}.
\begin{table*}[t]
    \centering
      \resizebox{0.95\textwidth}{!}{
        \begin{tabular}{lcccccccc}
          \toprule
          \multicolumn{1}{c}{\multirow{2}{*}{Model}}  & \multicolumn{2}{c}{In-Domain}
          & \multicolumn{4}{c}{Out-of-Domain}
          &              \\
          \cmidrule(lr){2-3} \cmidrule(lr){4-7}
          \multicolumn{1}{c}{}
           & MATH                  & GSM8K                 &
          College               & DM                    & Olympiad
          & Theorem               & AVG                   \\     
          \midrule
          Standard \#1 & 17.4 & 63.1 & 12.1 & 23.1 & 3.7 & 9.6 & 21.5 \\
          Standard \#2 & 17.6 & 63.7 & 12.6 & 20.6 & 4.3 & 8.9 & 21.3 \\
          Standard \#3 & 17.5 & 65.4 & 12.9 & 21.6 & 4.7 & 10.9 & 22.2 \\
          Standard (Avg.) & 17.5\textsubscript{$\pm$0.1} & 64.1\textsubscript{$\pm$1.2} & 12.5\textsubscript{$\pm$0.4} & 21.8\textsubscript{$\pm$1.3} & 4.2\textsubscript{$\pm$0.5} & 9.8\textsubscript{$\pm$1.0} & 21.7\textsubscript{$\pm$0.5} \\
          \midrule
          \method~\#1 & 45.6 & 79.9 & 27.1 & 44.4 & 17.2 & 19.5 & 39.0 \\
          \method~\#2 & 45.3 & 79.8 & 27.5 & 45.4 & 17.0 & 19.4 & 39.1 \\
          \method~\#3 & 46.5 & 79.2 & 27.9 & 43.4 & 17.2 & 20.0 & 39.0 \\
          \method~(Avg.) & 45.8\textsubscript{$\pm$0.6} & 79.6\textsubscript{$\pm$0.4} & 27.5\textsubscript{$\pm$0.4} & 44.4\textsubscript{$\pm$1.0} & 17.1\textsubscript{$\pm$0.1} & 19.6\textsubscript{$\pm$0.3} & 39.0\textsubscript{$\pm$0.1} \\
          \bottomrule
        \end{tabular}
      }
    \caption{
      Performance comparison between the standard setting and \method~accross six benchmarks with three random runs. The average performance is reported with the standard deviation.
    }
\label{tab:sig_test}
\end{table*}
    
\begin{table*}[t]
    \centering
        \resizebox{\textwidth}{!}{
        \begin{tabular}{lccccccccc}
          \toprule
          \multicolumn{1}{c}{\multirow{2}{*}{Model}}                &
          \multirow{2}{*}{\# Samples} & \multicolumn{2}{c}{In-Domain}
          & \multicolumn{4}{c}{Out-of-Domain}
          &              \\
          \cmidrule(lr){3-4} \cmidrule(lr){5-8}
          \multicolumn{1}{c}{}
          &      & MATH                  & GSM8K                 &
          College               & DM                    & Olympiad
          & Theorem               & AVG                   \\     \midrule
          Standard & 15K & 12.4 & 60.3 &  8.4 & 17.0 & 2.2 & 7.6 & 18.0\\
          GPT Rewritten & 15K & 20.1 & 70.3 & 9.1 & 13.9 & 2.8 & 8.9 & 20.9\\
          Mosaic-IT & 15K & 11.7 & 40.9 & 7.4 & 9.2 & 2.7 & 9.9 & 13.6 \\
          Mosaic-IT + Original GSM8K and MATH & 30K & 11.0 & 54.7 & 6.9 & 9.8 & 1.9 & 9.5 & 15.6 \\
          DART-Math & 60K & 34.1 & 77.2 & 23.4 & 36.0 & 8.7 & 18.2 & 32.9\\
          \midrule
          \method~(\textit{Sequential}) & 30K & 32.7 & 73.9 & 18.9 & 29.3 & 9.3 & 15.5 & 29.9\\
          \method~(\textit{Parallel}) & 30K & 30.9 & 75.1 & 20.9 & 26.5 & 11.0 & 15.2 & 29.9\\
          \method~(\textit{Conditional}) & 30K & 26.3 & 73.0 & 15.6 & 21.4 & 7.3 & 12.8 & 26.1\\
        \method~(DeepSeekMath-7B-RL) & 60K & 42.0 & 78.1 & 24.0 & 36.5 & 13.0 & 13.8 & 34.6 \\
          \method & 60K & 41.6 & 79.8 & 24.3 & 39.2 & 13.6 & 18.1 & 36.1 \\
          \bottomrule
        \end{tabular}
      }
      \caption{
        Additional results based on Mistral-7B.
      }
  \label{tab:ablation_gpt_rewrite}
\end{table*}

\subsection{Fused Error Analysis}
\label{sec:appendix_problem_analysis}
In practice, we find that some fused problems are unreasonable or ambiguous, which are shown in Section~\ref{sec:appendix_more_cases}.
The reason may be that some problems are not suitable for fusion or the limited capacity of the model for generating fused problems.
To verify the correctness of the fused problems and their influence on the model's performance, we conduct an error analysis on the fused problems.
Specifically, borrowing the idea from rejection sampling~\citep{yuan2023rft}, we use GPT-4o-mini to verify the correctness and completeness of the fused problems.
The corresponding evaluation prompt is shown in Section~\ref{sec:appendix_prompts}.
For each identified unreasonable problem, we adjust the temperature to 1.0 to enhance the diversity of generation, and re-generate the problems five times using the corresponding fusion strategy.
If none of the five generated problems is reasonable, we consider the fusion to be unreasonable and discard it.
Finally, 5.6\% of the fused problems are identified as unreasonable, and the remaining reasonable problems are added to the dataset.
The average performance of Llama3-8B fine-tuned only on the filtered \dataset~is 39.1, which is similar to the performance of the model fine-tuned on the original \dataset~(39.0), indicating that the unreasonable problems have little impact on the model's performance.
This result aligns with the findings in OpenMathInstruct-2~\citep{toshniwalopenmathinstruct-2}, indicating models exhibit some robustness to low-quality data in SFT.

\section{Effect of Teacher Model}
\label{sec:appendix_teacher_model}
In \method, we use GPT-4o-mini~\citep{openai2023gpt4} as the teacher model to generate the solutions for the fused problems.
To validate the performance improvement of \method~is not merely due to the stronger teacher model, we conduct two ablation studies:
(1) use GPT-4o-mini to rewrite the solutions from the original training set; and
(2) follow DART-Math~\citep{dartmath} to use DeepSeekMath-7B-RL~\citep{deepseekmath} to generate solutions for the fused problems.
The results are shown in Table~\ref{tab:ablation_gpt_rewrite}.
We can see that the performance of the model fine-tuned on the rewritten solutions is better than the Standard setting, especially on the MATH and GSM8K datasets.
However, the average improvement is only 2.9 points.
Meanwhile, each fusion strategy of \method~still outperforms the rewritten solution by a large margin.
Additionally, though DeepSeekMath-7B-RL underperforms GPT-4o-mini in distillation quality (34.6 v.s. 36.1 on average), it still outperforms DART-Math (34.6 v.s. 32.9 on average).
These results indicate that the performance improvement of \method~mainly comes from the new problem generated by three fusion strategies rather than the stronger teacher model.

\section{Additional Baseline}
A most recent work, Mosaic-IT~\citep{mosaic}, shares similar idea with our \method.
Mosaic-IT is a model-free data augmentation technique that operates by concatenating existing instruction-following datasets and subsequently training LLMs using these augmented data instances along with meta-instructions.
We conduct a comparison with the ``Primary Strategy'' proposed in Mosaic-IT, where the question pairs (same as \method) and corresponding solutions from the original GSM8K and MATH datasets are concatenated into a single sample for SFT, resulting in 15K data. 
To mitigate overfitting to the pattern of answering multiple questions jointly, we also conduct an additional experiment which combine Mosaic-IT and the original GSM8K and MATH training sets during training, resulting in 30K data in total. 
The results are shown in Table~\ref{tab:ablation_gpt_rewrite}. 
We observe that the Mosaic-IT leads to inferior performance, even worse than the Standard setting (i.e., using only the original GSM8K and MATH training data).
We suspect this may be due to the lack of logical integration between problems when simply concatenated—unlike \method, which explicitly introduces semantic or reasoning connections (e.g., sequential dependency or comparative logic) through its fusion strategies. 
This highlights the advantage of model-driven, structure-aware fusion over model-free concatenation.

\section{Significant Test}
We conduct error analysis on \method~on Llama3-8B model to verify the consistent performance improvement of our \dataset.
Specifically, we fine-tune the Llama3-8B model on the original training sets (Standard setting), and the combined fusion strategies, respectively.
The results are shown in Table~\ref{tab:sig_test}.
We can see that the \method~models consistently outperform the standard setting across all benchmarks.
We also conduct statistical significance tests using the paired t-test, and results show that the performance improvement of \method~is statistically significant ($p<0.05$) on all benchmarks.

\begin{table*}
\begin{trainprompt}{\textit{Sequential Fusion}}{seq_fusion}
\small
\textit{\# Role: Mathematical Problem Merger\\
\\
\#\# Profile\\
Your role is to merge "\#Problem 1\#" and "\#Problem 2\#" into a combined problem.\\
\\
\#\# Guidelines\\
Step 1: Identify input and output variables in both problems. Determine mathematical relationships and constraints in each problem. Locate variables between "\#Problem 1\#" and "\#Problem 2\#" that can form sequential dependencies.\\
\\
Step 2: Formulate a comprehensive plan to merge the two problems by using "\#Problem 1\#"’s output variable to replace an input variable of "\#Problem 2\#"’s. Merge contextual elements by embedding both problems within a unified real-world scenario or extended narrative, aligning units and measurement systems.\\
\\
Step 3: Create a single "\#Combined Problem\#" where solving "\#Problem 1\#" is a prerequisite for "\#Problem 2\#". Explicitly state variable dependencies and which variable is replaced. Adjust numerical ranges to maintain arithmetic consistency. The "\#Combined Problem\#" should contain no supplementary explanation or note.\\
\\
\#\# Output Format\\
Please reply strictly in the following format:\\
\#Elements Identified\#:\\
\#Plan\#:\\
\#Combined Problem\#:\\
\\
\#\# Input\\
\#\#\# \#Problem 1\#\\
\{problem1\}\\
\\
\#\#\# \#Problem 2\#\\
\{problem2\}\\
\\
\#\# Output
}
\end{trainprompt}
\end{table*}
\begin{table*}
\begin{trainprompt}{\textit{Parallel Fusion}}{para_fusion}
\small
\textit{\# Role: Mathematical Problem Synthesizer\\
\\
\#\# Profile
Your role is to organically integrate "\#Problem 1\#" and "\#Problem 2\#" to create a novel problem that requires advanced synthesis of their mathematical essence.\\
\\
\#\# Guidelines\\
Step 1: Conduct deep structural analysis of both problems by identifying their fundamental mathematical operations, contextual frameworks, and cognitive patterns. Extract the underlying logical architectures while preserving their distinctive solution pathways.\\
\\
Step 2: Develop an innovative fusion mechanism by discovering non-obvious mathematical connections between the problems' core concepts. Construct a multidimensional scenario that naturally embeds both original contexts through temporal sequencing, spatial superposition, or conceptual analogy. Engineer hybrid parameters that inherit characteristics from both source problems while introducing emergent properties.\\
\\
Step 3: Formulate the synthesized problem through strategic recombination of mathematical elements, ensuring the new problem requires concurrent application of both original solution strategies. Introduce controlled complexity through cross-domain constraints and self-verification mechanisms that establish mathematical consistency with both source problems' answers.\\
\\
\#\# Output Format\\
Please reply strictly in the following format:\\
\#Core Elements\#:\\
\#Synthesis Method\#:\\
\#New Problem\#:\\
\\
\#\# Input\\
\#\#\# \#Problem 1\#\\
\{problem1\}\\
\\
\#\#\# \#Problem 2\#\\
\{problem2\}\\
\\
\#\# Output
}
\end{trainprompt}
\end{table*}
\begin{table*}
\begin{trainprompt}{\textit{Conditional Fusion}}{cond_fusion}
\small
\textit{\# Role: Problem Integrator\\
\\
\#\# Profile\\
Create a real-world problem where the solution requires solving both "\#Problem 1\#" and "\#Problem 2\#" independently. **Ensure the the final answer is either from "\#Problem 1\#" or "\#Problem 2\#", depends on the "\#New Question\#"**.\\
\\
\#\# Guidelines\\
Step 1: Analyze "\#Problem 1\#" and "\#Problem 2\#" and make sure that the output variables they ask about are of the same type. If they are different (for example, one asks about time and the other asks about price), modify one of the problem so that it asks about the same variable as the other.
\\
Step 2: Design a unified problem scenario that combines "\#Problem 1\#" and "\#Problem 2\#". Introduce a "\#New Question\#", which must be related with both "\#Problem 1\#" and "\#Problem 2\#". Ensure that final answer of the "\#New Question\#" must either come from "\#Problem 1\#" or "\#Problem 2\#". This means that the "\#New Question\#" should be an **comparison** and **selection** of the previous answers, not their **combination**. There are some examples for the "\#New Question\#":
\begin{enumerate}
    \item Who sells the most items?
    \item How much money does the top earner make?
    \item Which is the cheaper plan?
    \item Someone has 200 dollor, which item can he afford?
\end{enumerate}
Step 3: Provide the "\#New Problem\#", which combine "\#Problem 1\#", "\#Problem 2\#", and "\#New Question\#" in a unified real-world scenario. Don't contain solution of "\#Problem 1\#" and "\#Problem 2\#" in "\#New Problem\#". Avoid using the phrases "\#Problem 1\#" and "\#Problem 2\#" in the generated "\#New Problem\#".\\
\\
\#\# Output Format\\
Please reply strictly in the following format:\\
\#Analysis\#:\\
\#New Question\#:\\
\#New Problem\#:\\
\\
\#\# Input\\
\#\#\# \#Problem 1\#\\
\{problem1\}\\
\\
\#\#\# \#Problem 2\#\\
\{problem2\}\\
\\
\#\# Output
}
\end{trainprompt}
\end{table*}
\begin{table*}
\begin{trainprompt}{Problem Evaluation}{problem_eval}
\small
\textit{\# Role: Mathematics Grading Teacher\\
\\
\#\# Profile\\
You are a senior mathematics grading teacher in university, very skilled in high difficulty fields such as Intermediate Algebra, Precalculus, Prealgebra, Number Theory, Geometry, Counting \& Probability, Algebra and so on.\\
\\
\#\# Guidelines\\
Your task is to act as an impartial judge to evaluate the statement completeness and correctness of math problem according to the following rules:
\begin{enumerate}
    \item Assess the clarity and accuracy of the definition of each math problem. Ensure that the problem statement provides sufficient information, conditions, and constraints.
    \item Consider whether the problem allows for multiple interpretations or if further clarification is needed.
    \item Evaluate the clarity of mathematical notation and terminology used in the problem.
    \item Evaluate whether the math problem is solvable.
    If the math problem meet the rules above, output "True" in "\#Judgement\#", else "False". You should also give your explanation in "\#Explanation\#".
\end{enumerate}
\#\# Output Format\\
Please reply strictly in the following format:\\
\#Judgement\#:\\
\#Explanation\#: \\
\\
\#\# Input\\
\{question\}\\
\\
\#\# Output
}
\end{trainprompt}
\end{table*}

\section{More Cases}
\label{sec:appendix_more_cases}
More cases, including the original problems \pa~and \pb, the fused problem \pf, are shown below.
Specifically, we show three reasonable cases in Case~\ref{exmp:case_1}, Case~\ref{exmp:case_2}, and Case~\ref{exmp:case_3}, and three unreasonable cases in Case~\ref{exmp:err_case_1}, Case~\ref{exmp:err_case_2}, and Case~\ref{exmp:err_case_3}.
\begin{table*}
    \begin{exmp}{Case 1}{case_1}
    \small
    \pa: Noah is a painter. He paints pictures and sells them at the park. He charges $\$60$ for a large painting and $\$30$ for a small painting. Last month he sold eight large paintings and four small paintings. If he sold twice as much this month, how much is his sales for this month?\\\\
    \pb: Michael loves to paint and sells his creations. He charges $\$100$ for a large painting and $\$80$ for a small painting. At his last art show, he sold 5 large paintings and 8 small paintings. How much did he earn in all?
    \\ \hdashrule{\linewidth}{0.5pt}{2pt} \\[0.4em]
    \pfseq: Noah is a painter who charges $\$60$ for a large painting and $\$30$ for a small painting. Last month, he sold 8 large paintings and 4 small paintings. This month, he sold twice as many paintings as last month. Calculate Noah's total sales for this month. Michael, another painter, charges $\$100$ for a large painting and $\$80$ for a small painting. \textcolor{red3}{At his last art show, he sold 5 large paintings and the same amount in dollars as Noah's total sales from this month in small paintings.} How much did Michael earn in all?
    \\ \hdashrule{\linewidth}{0.5pt}{2pt} \\[0.4em]
    \pfpara: \textcolor{red3}{Noah and Michael are collaborating in an art exhibit.} Noah charges $\$60$ for a large painting and $\$30$ for a small painting. Last month, he sold eight large paintings and four small paintings. This month, he plans to sell 50\% more of each type. Michael charges $\$100$ for a large painting and $\$80$ for a small painting. At his last show, he sold 5 large paintings and 8 small paintings. For the exhibit, he aims to sell 40\% more large paintings and 25\% more small paintings than last time. \textcolor{red3}{How much total sales will they collectively earn from the exhibit?}
    \\ \hdashrule{\linewidth}{0.5pt}{2pt} \\[0.4em]
    \pfcond: Noah is a painter who sells his artwork at the park. He charges $\$60$ for a large painting and $\$30$ for a small painting. Last month, he sold eight large paintings and four small paintings. This month, he has doubled his sales. Meanwhile, Michael, another painter, charges $\$100$ for a large painting and $\$80$ for a small painting. At his last art show, he sold 5 large paintings and 8 small paintings. \textcolor{red3}{Who earned more from their painting sales this month, Noah or Michael?}
\end{exmp}
\end{table*}
\begin{table*}
    \begin{exmp}{Case 2}{case_2}
    \small
    \pa: There were 133 people at a camp. There were 33 more boys than girls. How many girls were at the camp?\\\\
    \pb: At a party, there are 50 people. 30\% of them are boys. How many girls are at this party?
    \\ \hdashrule{\linewidth}{0.5pt}{2pt} \\[0.4em]
    \pfseq: At a camp, there were 133 people, with 33 more boys than girls. How many girls were at the camp? Subsequently, at a party where \textcolor{red3}{the total number of people is 50 plus the number of girls from the camp}, 30\% of them are boys. How many girls are at this party?
    \\ \hdashrule{\linewidth}{0.5pt}{2pt} \\[0.4em]
    \pfpara: At a camp, there are 133 people, and there are 33 more boys than girls. At a party, there are 50 people, and 30\% of them are boys. \textcolor{red3}{What is the total number of girls at both the camp and the party?}
    \\ \hdashrule{\linewidth}{0.5pt}{2pt} \\[0.4em]
    \pfcond: At a summer camp, there were 133 people. There were 33 more boys than girls. At the same time, at a party, there are 50 people, and 30\% of them are boys. \textcolor{red3}{How many girls are at each location, and which location has more girls?}
\end{exmp}
\end{table*}
\begin{table*}
\begin{exmp}{Case 3}{case_3}
\small
\pa: Add 53.463 to 12.9873 and round to the nearest thousandth.\\\\
\pb: Add $81.76$ to $34.587$ and round your answer to the nearest tenth.
\\ \hdashrule{\linewidth}{0.5pt}{2pt} \\[0.4em]
\pfseq: \textcolor{red3}{First}, add 53.463 to 12.9873 and round your answer to the nearest thousandth. \textcolor{red3}{Let this result be \( X \). Then, add \( X \) to 81.76 and 34.587}, and round your final answer to the nearest tenth.
\\ \hdashrule{\linewidth}{0.5pt}{2pt} \\[0.4em]
\pfpara: \textcolor{red3}{Calculate the sum of 53.463 and 81.76, then add this result to 34.587. Round the total to the nearest hundredth first, and then take that result and round it to the nearest whole number.} What is the final answer?
\\ \hdashrule{\linewidth}{0.5pt}{2pt} \\[0.4em]
\pfcond: A baker is preparing two different batches of cookies. For the first batch, she uses 53.463 ounces of flour and 12.9873 ounces of sugar. She wants to know the total weight of flour and sugar combined, rounded to the nearest thousandth. For the second batch, she uses $81.76 in ingredients and $34.587 in packaging. She needs to find the total cost of ingredients and packaging, rounded to the nearest tenth. \textcolor{red3}{After calculating both rounded totals, she wonders which amount is larger.}
\end{exmp}
\end{table*}

\begin{table*}
\begin{exmp}{Error Case 1}{err_case_1}
\small
\pa: What is the smallest positive integer with six positive odd integer divisors and twelve positive even integer divisors?\\\\
\pb: What is the smallest positive integer with exactly 10 positive integer divisors?
\\ \hdashrule{\linewidth}{0.5pt}{2pt} \\[0.4em]
\pfseq: Find the smallest positive integer \( x \) that has six positive odd integer divisors and twelve positive even integer divisors. Once \( x \) is determined, calculate the smallest positive integer \( y \) that has exactly 10 positive integer divisors.
\\ \hdashrule{\linewidth}{0.5pt}{2pt} \\[0.4em]
\textcolor{red3}{Error: The output of \pa~has no real relation to the input of \pb.}
\end{exmp}
\end{table*}
\begin{table*}
\begin{exmp}{Error Case 2}{err_case_2}
\small
\pa: In triangle $ABC$, $\cos(2A-B)+\sin(A+B)=2$ and $AB=4$. What is $BC$?\\\\
\pb: If triangle $ABC$ has sides of length $AB = 6,$ $AC = 5,$ and $BC = 4,$ then calculate
\[\frac{\cos \frac{A - B}{2}}{\sin \frac{C}{2}} - \frac{\sin \frac{A - B}{2}}{\cos \frac{C}{2}}.\]
\\ \hdashrule{\linewidth}{0.5pt}{2pt} \\[0.4em]
\pfpara: In triangle \( ABC \), the lengths of sides are \( AB = 4 \), \( AC = 5 \), and \( BC = 6 \). If it is given that \( \cos(2A - B) + \sin(A + B) = 2 \) holds true and the angles \( A, B, \) and \( C \) are interpreted through its trigonometric identities, compute the value of 
\[
\frac{\cos \frac{A - B}{2}}{\sin \frac{C}{2}} - \frac{\sin \frac{A - B}{2}}{\cos \frac{C}{2}}.
\]
Further, verify your answer by ensuring that the relationships between side lengths corroborate the angle calculations used to derive it.
\\ \hdashrule{\linewidth}{0.5pt}{2pt} \\[0.4em]
\textcolor{red3}{Error: The triangle can not satifsy both $AB = 6,$ $AC = 5,$ $BC = 4,$ and \( \cos(2A - B) + \sin(A + B) = 2 \).}
\end{exmp}
\end{table*}
\begin{table*}
\begin{exmp}{Error Case 3}{err_case_3}
\small
\pa: Andrew and John are both Beatles fans.  Their respective collections share nine of the same albums.  Andrew has seventeen albums in his collection.  Six albums are in John's collection, but not Andrew's.  How many albums are in either Andrew's or John's collection, but not both?\\\\
\pb: Andy, Beth, Charlie, and Daniel take a test with thirty questions. Andy and Beth together get the same number of questions wrong as Charlie and Daniel together. Andy and Daniel together get four more questions wrong than Beth and Charlie do together. If Charlie gets five questions wrong, how many questions does Andy get wrong?
\\ \hdashrule{\linewidth}{0.5pt}{2pt} \\[0.4em]
\pfcond: Andrew and John are both Beatles fans. Their respective collections share nine of the same albums. Andrew has seventeen albums in his collection, while six albums are in John's collection, but not Andrew's. Meanwhile, Andy, Beth, Charlie, and Daniel take a test with thirty questions. Andy and Beth together get the same number of questions wrong as Charlie and Daniel together. Andy and Daniel together get four more questions wrong than Beth and Charlie do together. If Charlie gets five questions wrong, how many unique albums are in either Andrew's or John's collection, but not both, compared to how many questions Andy got wrong?
\\ \hdashrule{\linewidth}{0.5pt}{2pt} \\[0.4em]
\textcolor{red3}{Error: There is no conditional relationship between the two problems.}
\end{exmp}
\end{table*}

\end{document}